%% file: main_rss.tex
\documentclass[conference]{IEEEtran}
\usepackage{times}

\usepackage[numbers]{natbib}
\usepackage{multicol}
\usepackage{multirow}

\usepackage[utf8]{inputenc} 
\usepackage[T1]{fontenc}    
\usepackage{url}            
\usepackage{booktabs}       
\usepackage{amsfonts}       
\usepackage{nicefrac}       
\usepackage{microtype}      
\usepackage{xspace}         
\usepackage{bm}
\usepackage{balance}
\usepackage{float}
\newcommand{\method}{DIAL\xspace}

\usepackage{listings}
\usepackage{xcolor}
\usepackage[normalem]{ulem}
\usepackage{soul}

\definecolor{Cerulean}{rgb}{0,0,0.95}
\definecolor{LimeGreen}{rgb}{0.15,0.65,0.15}
\definecolor{RoyalBlue}{rgb}{0.25,0.41,0.88}
\definecolor{Rose}{rgb}{1.0, 0.15, 0.21}
\definecolor{Orange}{rgb}{1.0, 0.5, 0.0}
\definecolor{Gray}{gray}{0.6}
\definecolor{Black}{gray}{0.0}
\definecolor{Purple}{rgb}{0.77,0.12,0.64}
\definecolor{codegreen}{rgb}{0,0.8,0}
\definecolor{codered}{rgb}{0.95,0,0.3}
\definecolor{codegray}{rgb}{0.5,0.5,0.5}
\definecolor{codepurple}{rgb}{0.58,0,0.82}
\definecolor{backcolour}{rgb}{0.95,0.95,0.95}
 \definecolor{lightgray}{gray}{0.9}
 
\lstdefinestyle{Python}{
    language        = Python,
    basicstyle      = \scriptsize\ttfamily,
    keywordstyle    = \color{black},
    keywordstyle    = [2] \color{black}, 
    stringstyle     = \color{black},
    commentstyle    = \color{blue}\ttfamily,
    backgroundcolor = \color{backcolour},
    breakatwhitespace=false,
    breaklines=true,
    basewidth=0.55em,
    tabsize=2
}

\usepackage{todonotes}
\usepackage{booktabs}
\usepackage{wrapfig}
\usepackage{xcolor}
\usepackage{caption}
\usepackage[numbers]{natbib}
\usepackage[colorlinks = true,
            linkcolor = black,
            urlcolor  = magenta,
            citecolor = blue,
            anchorcolor = blue]{hyperref}
\usepackage{url}

\definecolor{dataset_a}{HTML}{4285f4}
\definecolor{dataset_b}{HTML}{ffab40}
\definecolor{dataset_c}{HTML}{cc7d5d}

\input{math_commands.tex}

\title{Robotic Skill Acquisition via Instruction \\ Augmentation with Vision-Language Models}

\author{\authorblockN{
    Ted Xiao$^{1,*}$\thanks{$^*$ Equal contribution} \quad Harris Chan$^{1,2,*}$ \quad Pierre Sermanet$^1$ \quad Ayzaan Wahid$^1$ \quad Anthony Brohan$^1$ \\
    Karol Hausman$^1$ \quad Sergey Levine$^1$ \quad Jonathan Tompson$^1$}
    \authorblockA{$^1$Robotics at Google \quad $^2$University of Toronto}
    \authorblockA{Project website: \href{https://instructionaugmentation.github.io}{https://instructionaugmentation.github.io}}
}
\IEEEoverridecommandlockouts 

\begin{document}

\makeatletter
\let\@oldmaketitle\@maketitle%
\renewcommand{\@maketitle}{\@oldmaketitle%
    \centering
    \includegraphics[width=0.96\linewidth]{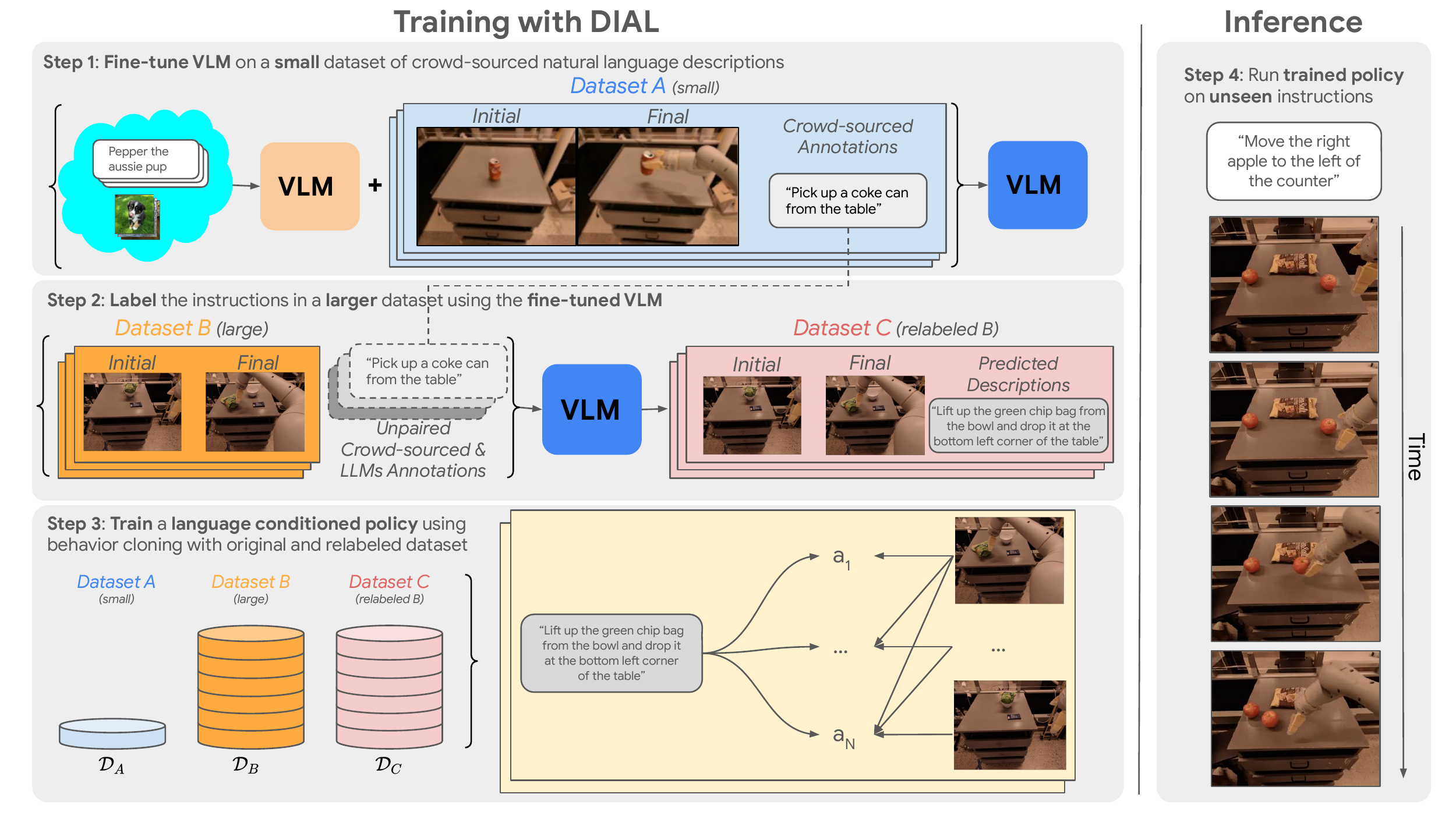}
    \captionof{figure}{\method consists of three steps: (1) Contrastive fine-tuning of a vision-language model (VLM) such as CLIP \citep{radford2021learning} on a small dataset of robot manipulation trajectories with crowd-sourced natural language annotation, (2) labeling a larger dataset of trajectories using the fine-tuned VLM (in dashed outline), and (3) training a language-conditioned policy using behavior cloning on the original and relabeled dataset. We evaluate the trained policy on unseen instructions. See Section \ref{sec:method} for more details.
    }
    \label{fig:main}
}
\makeatother
\maketitle

\begin{abstract}
Robotic manipulation policies that follow natural language instructions are typically trained from corpora of robot-language data that were either collected with specific tasks in mind or expensively relabeled by humans with varied language descriptions in hindsight. 
Recently, large-scale pretrained vision-language models (VLMs) have been applied to robotics for learning representations and scene descriptors.
Can these pretrained models serve as automatic labelers for robot data, effectively importing Internet-scale knowledge into existing datasets with limited ground truth annotations? 
For example, if the original annotations contained templated task descriptions such as ``pick apple'', a pretrained VLM-based labeler could significantly expand the number of semantic concepts available in the data and introduce spatial concepts such as ``the apple on the right side of the table'' or alternative phrasings such as ``the red colored fruit''.
To accomplish this, we introduce \linebreak
\textbf{D}ata-driven \textbf{I}nstruction \textbf{A}ugmentation for \textbf{L}anguage-conditioned control (\method): we utilize semi-supervised language labels to propagate CLIP's semantic knowledge onto large datasets of unlabeled demonstration data, from which we then train language-conditioned policies.
This method enables cheaper acquisition of useful language descriptions compared to expensive human labels, allowing for more efficient label coverage of large-scale datasets. 
We apply \method to a challenging real-world robotic manipulation domain where only 3.5\% of the 80,000 demonstrations contain crowd-sourced language annotations.
Through a large-scale study of over 1,300 real world evaluations, we find that DIAL enables imitation learning policies to acquire new capabilities and generalize to 60 novel instructions unseen in the original dataset.
\end{abstract}

\input{1_introduction}
\input{2_related_work}
\input{3_method}
\input{4_experimental_setup}
\input{5_experimental_results}
\input{6_conclusion}

\section*{Acknowledgement}
The authors would like to thank Kanishka Rao, Debidatta Dwibedi, Pete Florence, Yevgen Chebotar, Fei Xia, and Corey Lynch for valuable feedback and discussions.
We would also like to thank Emily Perez, Dee M, Clayton Tan, Jaspiar Singh, Jornell Quiambao, and Noah Brown for navigating the ever-changing challenges of data collection and robot policy evaluation at scale.
Additionally, Tom Small designed informative animations to visualize DIAL. 
Finally, we would like to thank the large team that built \citep{brohan2022rt} and \citep{ahn2022saycan}, upon which we develop DIAL.

\bibliographystyle{plainnat}
\bibliography{refs}

\clearpage
\appendix
\section{Appendix}
\input{7_appendix}

\end{document}

%% file: math_commands.tex
\usepackage{amsmath,amsfonts,bm}

\def\eqref#1{equation~\ref{#1}}

\def\1{\bm{1}}

\DeclareMathAlphabet{\mathsfit}{\encodingdefault}{\sfdefault}{m}{sl}
\SetMathAlphabet{\mathsfit}{bold}{\encodingdefault}{\sfdefault}{bx}{n}

\def\sR{{\mathbb{R}}}

\newcommand{\Ls}{\mathcal{L}}

\newcommand\norm[1]{\left\lVert#1\right\rVert}

%% file: 1_introduction.tex
\section{Introduction}
\label{sec:introduction}
Advances in deep learning architectures have made it possible to train end-to-end robotic control policies for following a wide range of textual instructions, often by integrating state-of-the-art language embeddings and pretrained encoders with imitation learning on top of large, manually collected datasets of robotic demonstrations annotated with text commands~\citep{jang2022bc}.
However the performance of such methods is critically dependent on the quantity and breadth of instruction-labeled demonstration data that is available~\citep{lynch2022interactive}, and producing expert demonstrations of robot motion often requires expertise and time~\citep{mandlekar2021matters}.
Can we squeeze more generalization capacity out of a given set of demonstrations, with minimal additional human effort?
The key observation we make is that a given demonstration might illustrate more than one behavior -- e.g., a motion that picks up the leftmost can in Figure 1 is an example of how to pick up a coke can, the left can in a row of three, a first step toward clearing the table, and more.
Can we leverage this observation to relabel a given demonstration dataset to enable a robot to master a broader range of semantic behaviors, and can we do this in a largely automated and scalable way?

One possibility is to leverage large-scale pretrained language models (LLMs)~\citep{brown2020language,devlin2018bert} and vision-language models (VLMs)~\citep{alayrac2022flamingo,radford2021learning}, which can be pretrained on Internet-scale data and then be applied to downstream domains.
In robotics, they have been used as representations for perception~\citep{nair2022r3m, shridhar2022cliport}, as task representation for language~\cite{jang2022bc, lynch2020language}, or as planners~\citep{ahn2022saycan, huang2022language}.
In contrast, we seek to apply pretrained VLMs to the datasets themselves: can we use VLMs for \textit{instruction augmentation}, where we relabel existing offline trajectory datasets with additional language instructions?

In this work, we introduce \textbf{D}ata-driven \textbf{I}nstruction \textbf{A}ugmentation for \textbf{L}anguage-conditioned Control (\method), a method that performs instruction augmentation with pretrained VLMs to weakly relabel offline control datasets.
We implement an instantiation of our method with CLIP~\citep{radford2021learning} on a challenging real-world robotic manipulation setting with 80,000 teleoperated demonstrations, which include 2,800 demonstrations that are labeled by crowd-sourced language annotators.
By performing a large quantitative evaluation of over 1,300 real world robot evaluations, we compare our method with baselines and instruction augmentation methods that are not visually grounded.
We find that DIAL enables policies to acquire understanding of new concepts not contained in the original task labels and improving performance on 60 novel evaluation instructions by over 41\%.
Sample emergent capabilities of our method are shown in Figure~\ref{fig:qualitative_results}.

%% file: 2_related_work.tex
\section{Related Work}
\label{sec:rw}
\paragraph{Language instruction following in robotics}
Language-instruction following agents have been extensively explored with engineered symbolic representations \citep{duvallet2013imitation,tellex2011approaching}, with reinforcement learning (RL) \citep{bahdanau2018learning,fu2019language,luketina2019survey}, and with imitation learning \citep{anderson2018vision,blukis2019learning,jang2022bc,lynch2020language}.
Recent advances in deep learning with large amounts of data have led to advances in methods for learning instruction-conditioned policies~\citep{ahn2022saycan,liang2022code,mees2022grounding,silva2021lancon,stepputtis2020language}. Latent Motor Policies (LMP) \citep{lynch2020learning} learns hierarchical goal-conditioned policies. Subsequent Language from Play (LfP) \citep{lynch2020language} uses language goals provided by large dataset of hindsight human labels on robotic play data.  
Similarly, Interactive Language \citep{lynch2022interactive} uses crowd-sourced hindsight labels on diverse demonstration data for table-top object rearrangements. 
In contrast, our method does not rely on crowd-sourced language labels at scale, but instead leverages a modest number of language labels by using a learned model to provide weak hindsight labeling for the rest of the data. 

\paragraph{Pretrained VLMs and LLMs for language-conditioned control}
Prior works have leveraged pretrained VLMs and LLMs for language-conditioned control, as part of reward modeling \citep{fan2022minedojo,nair2022learning}, as part of the agent architecture \citep{nair2022r3m,shridhar2022cliport}, or as planners for long-horizon tasks \citep{ahn2022saycan,huang2022language,huang2022inner}.
MineCLIP \citep{fan2022minedojo} fine-tunes CLIP \citep{radford2021learning} encoders using a contrastive loss on a large offline dataset of Minecraft videos and optimizes a language-conditioned control policy on top of the finetuned CLIP representations through online RL.
LOReL \citep{nair2022learning} learns a reward function from offline robot datasets with crowd sourced annotations using a neural network trained from scratch combined with a pretraind DistilBERT sentence embedding \citep{sanh2019distilbert} using a binary cross entropy loss.
CLIPort \citep{shridhar2022cliport} uses a frozen CLIP vision and text encoders in combination with Transporter networks \citep{zeng2021transporter} for imitation learning. 
R3M \citep{nair2022r3m} uses representations pretrained constrastively on Ego4D \citep{grauman2022ego4d} human video datasets for robotic policy learning via imitation learning. 
For long-horizon language instructions, LLMs have been used as planners both in simulated \citep{huang2022language} and real-world robotics settings \citep{ahn2022saycan}.
Our approach fine-tunes CLIP on our \textit{real} robot offline dataset and is used for instruction augmentation for a behavior cloning agent, instead of directly using the CLIP model as a reward model and optimizing an RL agent.

\paragraph{Hindsight relabeling for goal-conditioned reinforcement learning}
The relabeling approach for goal-conditioned reinforcement learning \citep{plappert2018multi} is used in tabular \citep{kaelbling1993learning} and continuous \citep{andrychowicz2017hindsight} settings, where the desired goals are relabeled with achieved goals to generate positive examples in sparse reward environments. This method has been applied to environments with goals represented as images \citep{chebotar2021actionable}, task IDs \citep{kalashnikov2021mt}, and language instructions \citep{chan2019actrce,cideron2020higher,jiang2019language}. Previous works with language goals used environment simulators \citep{chan2019actrce,jiang2019language} or learned models \citep{cideron2020higher,roder2022grounding} to provide hindsight labels.
Our work introduces the novel contribution of visual grounding by leveraging VLMs to generate unstructured natural language relabeling instructions, enabling scaling to complex real robot environments.

%% file: 3_method.tex
\section{Data-driven Instruction Augmentation for Language-conditioned Control}
\label{sec:method}

In this section, we describe our method, DIAL, which consists of three stages:  (1) fine-tuning a VLM's vision and language representations on a small offline dataset of trajectories
with crowd-sourced episode-level natural language descriptions, 
(2) generating alternative instructions for a larger offline dataset of trajectories with the VLM, and (3) learning a language-conditioned policy via behavioral cloning on this instruction-augmented dataset. 

\subsection{Fine-tuning Vision-Language Model Representations}
\label{subsec:finetuning}
We first collect a dataset of robot trajectories, from either human teleoperated demonstrations on a wide variety of tasks \citep{ahn2022saycan}, or from unstructured robotic ``play'' data \citep{lynch2020learning}.
We partition this dataset with uniform sampling into two subsets: a small subset to be annotated by human annotators, and a much larger subset to be labeled by the VLM finetuned on the former. The smaller portion is selected because the process of human labeling is time-consuming and requires significant effort and cost.

Let the small dataset of $N$ trajectories be $[\tau_1,\dots,\tau_N]$, $\tau_n = ([(s^n_0, a^n_0),(s^n_1,a^n_1),\dots,(s^n_T)])$, where $s_t^n$ and $a_t^n$ denote the observed state and action, respectively, at time $t$ for the $n$-th episode. We then collect a corresponding natural language annotation $l^n$ for the $n$-th episode describing what the robot agent did in the episode via crowd-sourcing. 

When producing these descriptions, the crowd-sourced evaluators observe the first frame, $s_0$, and last frame, $s_T$, from the agent's first-person view. We refer to these instructions as \textit{crowd-sourced instructions}.
Together, we denote the first dataset $\mathcal{D}_A = [(\tau_1, l^1),\dots,(\tau_N, l^N)]$ as the paired trajectories and crowd-sourced labels.  
Our method then fine-tunes a vision and language model representation on $\mathcal{D}_A$.

Motivated by promising results of CLIP in robotics in prior works \citep{mees2022matters,shridhar2022cliport}, our instantiation of DIAL uses CLIP \citep{radford2021learning} for both instruction augmentation and task representation; nonetheless, other VLMs or captioning models could also be used to propose instruction augmentations.
Given a batch of $B$ initial state $s_0$, final state $s_T$, and crowd-sourced instruction $l$ tuple, the model is trained to predict which of the $B^2$ (initial-final state, crowd-sourced instruction) pairs co-occurred.
We use CLIP's Transformer-based text encoder $T_{enc}$ to embed the crowd-sourced instruction to a latent space $z^n_{l} = T_{enc}(l^n) / \norm{T_{enc}(l^n)} \in \sR^d$ and CLIP's Vision Transformer-based (ViT) \citep{dosovitskiy2020image} image encoder $I_{enc}$ to embed the initial and final state, and further concatenate these two embeddings and pass through a fully connected neural network $f_\theta$, producing the vision embedding $z^n_{s} = f_\theta([I_{enc}(s^n_0);I_{enc}(s^n_T)]) / \norm{f_\theta([I_{enc}(s^n_0);I_{enc}(s^n_T)])} \in \sR^d$. $B^2$ similarity logits are formed by applying dot product across all state-instruction pairs, and a symmetric cross entropy loss term is calculated by applying softmax normalization with temperature $\alpha$ across the states and texts:
\begin{align}
    \nonumber
    \Ls_\theta = - \sum_{n=1}^B \bigg[ \log\bigg(\frac{e^{z^n_{l} \cdot z^n_{s} / \alpha}}{\sum_{k=1}^B e^{z^k_{l} \cdot z^n_{s}  / \alpha}}\bigg) 
     + \log\bigg(\frac{e^{z^n_{l} \cdot z^n_{s} / \alpha}}{\sum_{k=1}^B e^{z^n_{l} \cdot z^k_{s} / \alpha}}\bigg) \bigg]
\end{align}

\subsection{Instruction Augmentation}
\label{subsec:relabeling}
In the larger partition of the original dataset, which we denote as dataset $\mathcal{D}_B$, contains $M \gg N$ trajectories $[\hat{\tau}_1,\dots,\hat{\tau}_{M}]$, where  $\hat{\tau}_m = ([(\hat{s}^m_0, \hat{a}^m_0),(\hat{s}^m_1,\hat{a}^m_1),\dots,(\hat{s}^m_T)])$.
In contrast to $\mathcal{D}_A$, we assume that trajectories in $\mathcal{D}_B$ do not have any associated natural language labels.
\begin{figure}[t]
    \begin{center}
        \includegraphics[width=
        0.8\linewidth]{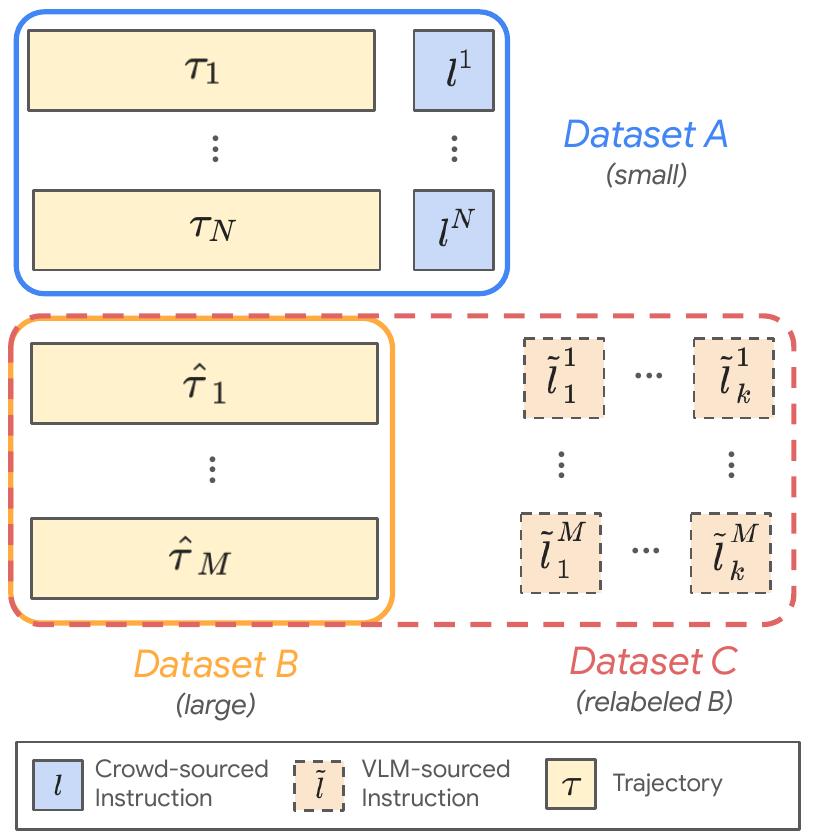}
    \end{center}
    \caption{The construction of datasets: Dataset A  ($\mathcal{D}_A$) (\textcolor{dataset_a}{blue}) consists of the $N$ trajectories ${\{\tau_n\}^N_{n=1}}$ labeled with crowd-sourced instructions ${\{l^n\}^N_{n=1}}$ describing what the robot agent performed in the episode. 
    Dataset B ($\mathcal{D}_B$) (\textcolor{dataset_b}{yellow}) consists of a much larger set of trajectories, $\{\hat{\tau}_m\}^M_{m=1}$
    \textit{without} crowd-sourced instructions. 
    Dataset C ($\mathcal{D}_C$) (\textcolor{dataset_c}{red, dashed}) contains Dataset B trajectories relabeled with VLM-sourced hindsight instruction(s) ${\{\tilde{l}^m_1,\dots,\tilde{l}^m_k\}^M_{m=1}}$.
    }
    \label{fig:datasets}
    \vspace{-12pt}
\end{figure}

\begin{figure*}[t]
    \centering
    \includegraphics[width=0.95\linewidth]{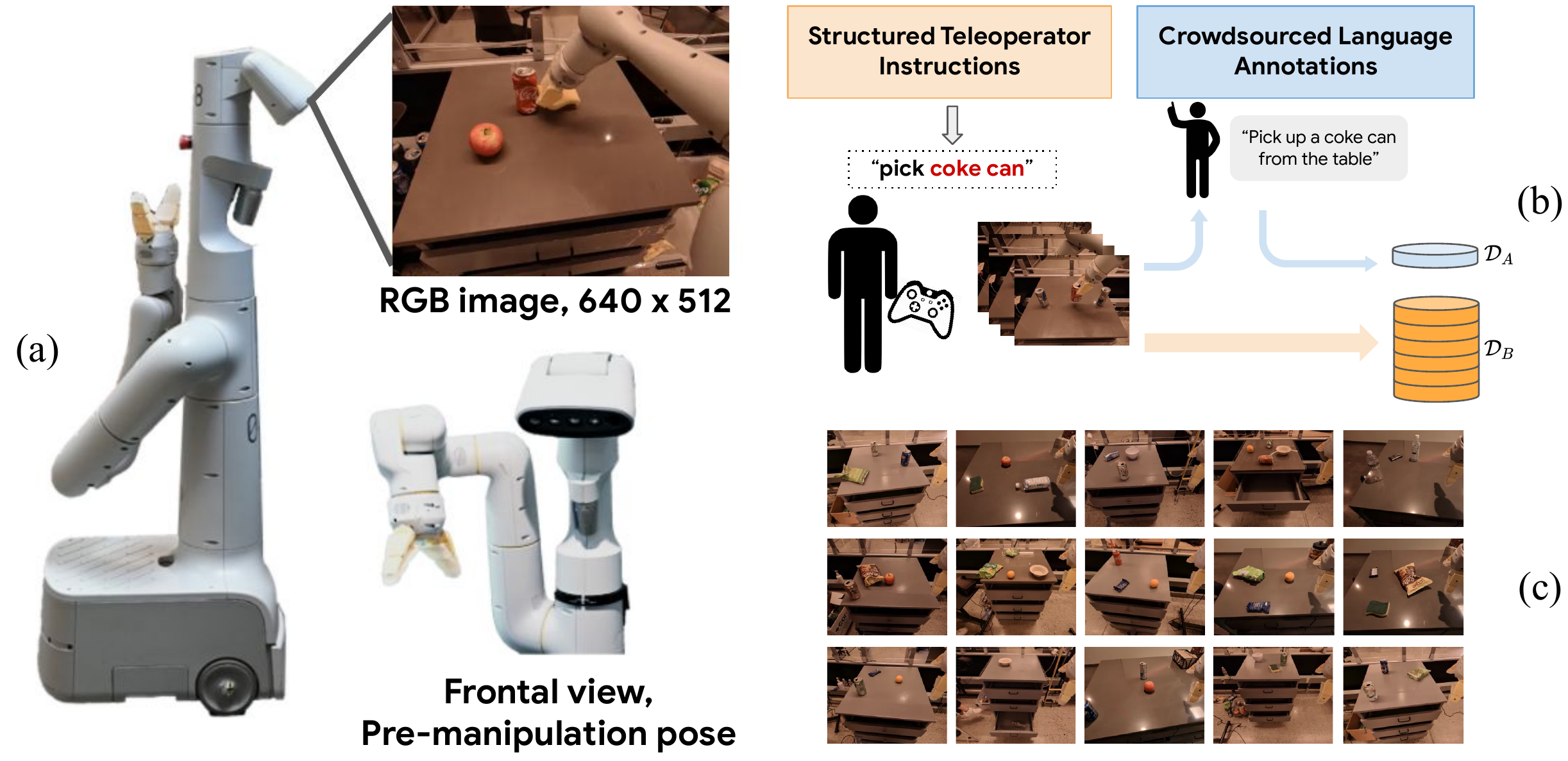}
    \caption{(a) A mobile manipulator robot receives RGB images from an onboard camera and uses a 7 DoF arm with parallel-jaw grippers. (b) Teleoperators receive instructions drawn from a set of 551 structured commands to perform a total of 80,000 demonstrations. 2,800 of these episodes are sent for crowdsourced language annotations. (c) A sample of scenes in the demonstration dataset which range across various countertops, drawers, and object arrangements in an office kitchen setting.}
    \label{fig:setup}
\end{figure*}

We use the fine-tuned VLM model to propose natural language instructions $\tilde{l}^m$ for each \mbox{trajectory} $\hat{\tau}_m$ to augment $\mathcal{D}_B$.
While $\tilde{l}^m$ could be drawn from any reasonable corpus, our \mbox{specific} instantiation of DIAL sources these \mbox{candidate} instructions from $\mathcal{D}_A$ as well as \mbox{additional} instructions drawing from GPT-3~\citep{brown2020language} \mbox{proposals} of possible tasks, which we denote as $\mathcal{D}_{GPT-3}$ (the details of this procedure will be covered in Section~\ref{subsec:instruction_augmentation}).
We use the CLIP text encoder to \mbox{independently} embed these candidate natural language instructions, i.e. $\tilde{l}^m \in L = \{l^1,\dots,l^N\} \sim \mathcal{D}_A \cup \mathcal{D}_{GPT-3}$:
\begin{align*}
    \{z^1_{l},\dots,z^N_{l}\} = \{T_{enc}(l^1),\dots,T_{enc}(l^N)\}
\end{align*} 
Similarly, we use the fine-tuned CLIP image encoder and MLP fusion to embed the initial and final observations from the second dataset: 
\begin{align*}
    \{\hat{z}^1_{s},\dots,\hat{z}^M_{s}\} = \{f_\theta([I_{enc}(\hat{s}^i_0);I_{enc}(\hat{s}^i_T)])\}_{i=1}^M
\end{align*} 

With these embeddings pre-computed, we can retrieve the most likely candidates using $k$-Nearest Neighbors \citep{fix1989discriminatory} with cosine similarity between the vision-language embedding pairs  $d(z^n_l, \hat{z}^m_s) = \frac{z^n_l \cdot \hat{z}^m_s}{\norm{z^n_l \cdot \hat{z}^m_s}}$ as the metric. We then use the cosine similarity to select a subset of candidate instructions to construct a new \textit{relabeled} dataset 
$\mathcal{D}_C = [(\hat{\tau}_1,\tilde{l}^1_1),\dots,(\hat{\tau}_1,\tilde{l}^1_k),\dots, (\hat{\tau}_M, \tilde{l}^M_1), \dots, (\hat{\tau}_M,\tilde{l}^M_k)]$. Figure \ref{fig:datasets} visualizes the three datasets generated.
There are several potential strategies for candidate instruction selection:
\paragraph{Top-$k$ selection} 
For each trajectory, we rank the candidate instructions in descending order based on their cosine similarity distances and output the top-$k$ instructions. 
The hyperparameter $k$ trades off precision and recall of the relabeled dataset. A smaller $k$ will return mostly relevant candidate instructions, while a larger $k$ value can recall a broader spectrum of potential hindsight descriptions for the episode at the expense of introducing erroneous instructions.  

\paragraph{Min-$p$ selection}
Instead of outputting a fixed number of candidate instructions per trajectory, we dynamically adjust this number based on a minimum probability $p$ parameter, representing the minimum confidence for each instruction.
We first convert the cosine similarity between the vision-language embedding pair to a probability that the $m$-th episode has language label $l^n$ by taking the softmax over all the candidate instructions with temperature parameter $\alpha$ from CLIP:
\begin{align}
    P(\tilde{l}^m = l^n | (\hat{s}^m_0, \hat{s}^m_T)) = \frac{\exp(d(z^n_l, \hat{z}^m_s)/\alpha)}{\sum_{n'} \exp(d(z^{n'}_l, \hat{z}^m_s)/\alpha)}
\end{align}
We truncate the candidate instructions to the set $L^{(p)} \subset L$ such that each instruction has a minimum hurdle probability $p > 0$:
\begin{align}
    P(\tilde{l}^m = l) \geq p, \quad \forall \ l \in L^{(p)}
\end{align}

Given $p$, the \textit{maximum} number of candidates that can be output for a trajectory is $k= 1/p$. The \textit{minimum} number of candidates, meanwhile, can be zero, if there are no candidate instructions satisfying this hurdle probability. 

We will investigate in Section \ref{subsec:accuracy_tradeoff} the effects of these candidate instruction selection strategies on relabeled instruction accuracy, augmented dataset size, and downstream policy performance. 

\subsection{Language Conditioned Policies with Behaviour Cloning}
\label{subsec:lang_bc}
Given a dataset $\mathcal{D} = [\mathcal{D}_A, \mathcal{D}_C]$ of robot trajectories and corresponding augmented language instructions, we can train a language-conditioned control policy with Behavior Cloning (BC).
While instruction augmented offline datasets can be used by any downstream language-conditioned policy learning method such as offline RL or BC, we limit our work to the conceptually simpler BC in order to focus our analysis on the importance of instruction augmentation.

%% file: 4_experimental_setup.tex
\section{Experimental Setup}
\label{sec:exp_setup}

We first describe the setup for our experimental validation, including the environments, the configuration of the robot, and the datasets that we use in our experiments. Since our aim is to study the benefits of our proposed instruction augmentation approach, we describe other alternative methods for augmenting the dataset for comparison. Finally, we detail our evaluation protocol, which involves testing the degree to which policies learned with different types of instruction augmentation generalize to \emph{novel previously unseen} instructions.

\subsection{Environment, Robot, and Datasets}
\label{subsec:environment}
We implement \method in a challenging real-world robotic manipulation setting based on the kitchen environments described by \citet{ahn2022saycan}.
We focus on the practically-motivated setting where a dataset of teleoperated demonstrations is available, collected for downstream imitation learning~\citep{ahn2022saycan, jang2022bc}.
A mobile manipulator robot with a parallel-jaw gripper, an over-the-shoulder RGB camera, and a 7 DoF arm is placed in an office kitchen to interact with common objects using concurrent~\citep{xiao2020nonblocking} continuous closed-loop control from pixels.
We collect a large-scale dataset of over 80,000 robot trajectories via human teleoperation ($\mathcal{D}_B$), where teleoperators receive 551 structured commands motivated by common manipulation skills and objects in a kitchen environment, following prior work~\citep{ahn2022saycan}.
Afterwards, we leverage crowd-sourced human annotators to label 2,800 robot trajectories with two hindsight instructions each, resulting in a total of 5,600 unique episodes with crowdsourced captions ($\mathcal{D}_A$).
Human annotators are shown the first and last frame of the episode and asked to provide a free-form text description describing how a robot should be commanded to go from the start to the end.
Visualizations of the mobile manipulator, dataset collection procedure, and example scenes are shown in Figure \ref{fig:setup} and further detailed in Appendix~\ref{app:datasets}.

\subsection{Instruction Augmentation}
\label{subsec:instruction_augmentation}

We consider various methods of instruction augmentation which each result in different relabeled datasets that are then used for downstream policy learning.

\paragraph{DIAL implementations} 
We implement DIAL with a CLIP model that is fine-tuned on $\mathcal{D}_A$ with the procedure described in Section~\ref{subsec:finetuning}.
After fine-tuning CLIP, we source 18,719 candidate instruction labels ($L$) from the combination of $\mathcal{D}_A$ and a corpus of GPT-3 proposals of potential language instructions.
To perform instruction augmentation that relabels dataset $\mathcal{D}_B$ of 80,000 robot trajectories that do not contain crowd-sourced annotations with $L$, we follow Section~\ref{subsec:relabeling} to implement two variations of DIAL: \textbf{Top-$k$ selection} and \textbf{Min-$p$ selection}.

The version of DIAL with \textbf{Top-$k$ selection} applies a fixed number $k$ instruction augmentations for every episode in the source dataset based on cosine similarity distances.
By changing \textit{k}, we produce three instruction augmented datasets: 80,000 relabeled demonstrations (\textit{k} = 1), 240,000 relabeled demonstrations (\textit{k} = 3), and 800,000 relabeled demonstrations (\textit{k} = 10).
The version of DIAL with \textbf{Min-$p$ selection} is more conservative and only performs instruction augmentation when confidence from CLIP is above some threshold $p$.
By changing $p$, we produce three instruction augmented datasets: 128,422 relabeled demonstrations (\textit{p} = 0.1), 38,516 relabeled demonstrations (\textit{p} = 0.2), and 17,013 relabeled demonstrations (\textit{p} = 0.3).
Additional details can be found in Appendix~\ref{app:dial_implementation}.

\paragraph{Non-visual instruction augmentation methods}
We consider three instruction augmentation methods that do \textit{not} utilize any visual information.
First, we implement a ``Gaussian Noise'' baseline that adds random noise to existing crowd-sourced instructions' language embeddings.
Second, we design a ``Word-level Synonyms'' baseline that replaces individual words in existing instructions with sampled synonyms from a predefined list.
Finally, we introduce a ``LLM-proposed Instructions'' baseline that replaces entire instructions with alternative instructions as proposed by GPT-3.
Implementation details for these baselines can be found in Appendix~\ref{app:baseline_implementations}.

\subsection{Policy Training}
\label{subsec:policy_training}
Using these various instruction augmented datasets, we train vision-based language-conditioned behavioral cloning policies with the RT-1 architecture~\citep{rt12022arxiv}. 
One main difference from RT-1 is that instead of utilizing USE~\citep{cer2018universal} as the task representation for language conditioning, we instead use the language encoder of the fine-tuned CLIP model that was used for  instruction augmentation in Section~\ref{subsec:relabeling}; full details are described further in Appendix~\ref{app:lcbc}. 
Nonetheless, we treat the behavioral cloning policy as an independent component of our method and focus on studying instruction augmentation methods; we do not explore different policy architectures or losses in this work.

\subsection{Evaluation}
\label{subsec:evaluation}

\begin{table}[t]
\small
\small
\begin{center}
\renewcommand{\arraystretch}{1.2}
         \centering
\begin{tabular}{|p{0.09\textwidth}|p{0.35\textwidth}|}
\hline
\textbf{Category} & \textbf{Instruction Samples}\\ \hline
\textit{Spatial} & \texttt{[`knock down the right soda', `raise the left most can', `raise bottle which is to the left of the can']} \\ \hline
\textit{Rephrased} & \texttt{ [`pick up the apple fruit', `liftt the fruit' [sic], `lift the yellow rectangle']} \\ \hline
\textit{Semantic} & \texttt{[`move the lonely object to the others', `push blue chip bag to the left side of the table', `move the green bag away from the others']} \\ \hline
\end{tabular}
\caption{Samples from the 60 novel evaluation instructions we consider.
34 \textit{Spatial} tasks focus on instructions involving reasoning about spatial relationships, such as specifying an object's initial position relative to other objects in the scene.
16 \textit{Rephrased} tasks are linguistic re-phrasings of the original 551 foresight tasks, such as referring to sodas and chips by their colors instead of their brand name.
10 \textit{Semantic} tasks describe skills not contained in the original dataset, such as moving objects away from all other objects, since the original dataset only contains trajectories of moving objects towards other objects.
A full list is provided in Table~\ref{tab:full_eval_tasks}.}
\label{tab:eval_categories}
\end{center}
\end{table}

\begin{figure*}[t]
    \centering
    \includegraphics[width=0.9\linewidth]{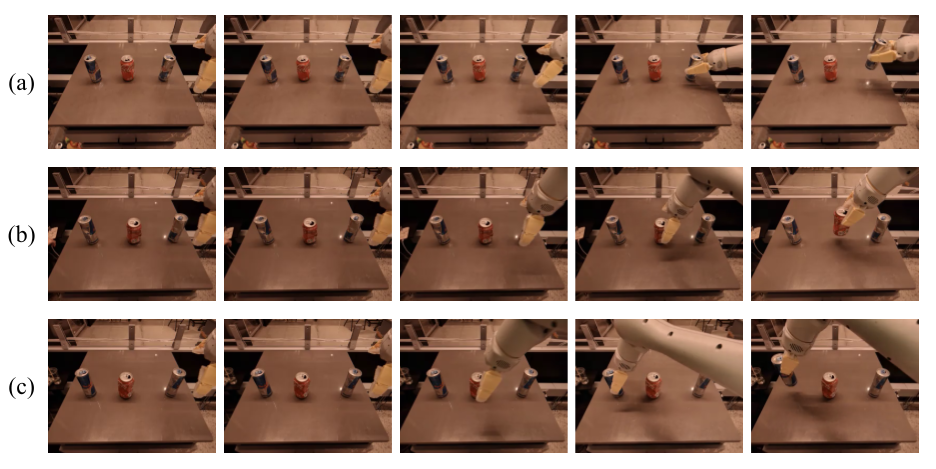}
    \caption{Given the same starting scene, DIAL follows the instructions of (a) \texttt{pick can which is on the right of the table}, (b) \texttt{pick the can in the middle}, and (c) \texttt{pick can which is on the left of the table}. Non-DIAL methods do not adjust their behaviors based on language commands, demonstrating a lack of spatial understanding.
    }
    \label{fig:qualitative_results}
\end{figure*}

In contrast to many prior works~\citep{anderson2018vision,blukis2019learning}
on instruction following, we focus our evaluation only on \textit{novel instructions unseen during training}.
To source these novel instructions, we crowd-source instructions and prompt GPT-3 for evaluation task suggestions, and then filter out any instructions already contained in either the crowd-sourced language instructions in $\mathcal{D}_A$, the original set of 551 structured teleoperator commands $\mathcal{D}_B$, or the instruction augmentated dataset $\mathcal{D}_C$; in total, we sample 60 novel evaluation instructions.
We organize these evaluation instructions into three categories to allow for more detailed analysis of qualitative policy performance; examples are shown in Table~\ref{tab:eval_categories} and a full list is provided in Table~\ref{tab:full_eval_tasks}.

%% file: 5_experimental_results.tex
\section{Experimental Results}
\label{sec:experiment_results}

In our experiments, we investigate whether DIAL can improve the policy performance on the unseen tasks described in Section \ref{subsec:evaluation} when starting from fully or partially labeled source datasets.
We ablate on the types of instruction augmentations described in Section \ref{subsec:instruction_augmentation}, and analyze the importance of accuracy when augmenting instructions with DIAL.  

\subsection{Does DIAL improve performance on unseen tasks?}
\label{subsec:exp1}
We investigate whether \method can enable language-conditioned behavior cloning policies to successfully perform novel instructions.
For training various control policies, we consider three training datasets: $\mathcal{D}_A$ contains 5,600 episodes which contain crowd-sourced hindsight language instructions, $\mathcal{D}_B$ contains 80,000 episodes which contain structured commands given to teleoperators, and $\mathcal{D}_C$ contains 38,516 episodes with instructions predicted by \method with Min-$p=0.2$ starting from $\mathcal{D}_A$ and $\mathcal{D}_B$.
We refer to training on only $\mathcal{D}_A$ as the Interactive Language (IL) \citep{lynch2022interactive} setting, training on only $\mathcal{D}_B$ as the RT-1 \citep{rt12022arxiv} setting, and training on both $\mathcal{D}_A$ and $\mathcal{D}_B$ as the RT-1 + IL setting.
Then, we refer training on $\mathcal{D}_C$ (either with or without $\mathcal{D}_A$ and $\mathcal{D}_B$) as \method.
This experiment is practically motivated by the setting where large amounts of unstructured trajectory data are available but hindsight labels are expensive to collect, such as robot play data ~\citep{cui2022play,lynch2020learning, lynch2022interactive}.

After policy training, we evaluate on task instructions not contained in $\mathcal{D}_A$, $\mathcal{D}_B$, or $\mathcal{D}_C$.
Table \ref{tab:partial_labels} demonstrates that DIAL is able to solve over 40\% more challenging novel tasks across the three evaluation categories compared to either RT-1 and/or IL, which do not use the instruction augmented data $\mathcal{D}_C$.
\begin{table}[t]
\begin{center}
\begin{tabular}{@{} c c c c p{0.7cm} p{0.9cm} p{0.7cm} p{0.5cm}@{}} 
 \toprule
  \multicolumn{1}{c}{\textbf{}} & \multicolumn{3}{c}{\textbf{Dataset Properties}} & \multicolumn{4}{c}{\textbf{Evaluation on Novel Instructions}} \\
  \cmidrule(lr){2-4} \cmidrule(lr){5-8}
 \shortstack{Method} & $\mathcal{D}_A$ & $\mathcal{D}_B$ & $\mathcal{D}_C$ & \shortstack{\textit{Spatial}} & \shortstack{\textit{Rephrased}} & \shortstack{\textit{Semantic}} & Overall \\ [0.5ex] 
 \midrule
 IL \citep{lynch2022interactive}  & \checkmark &  &  &$30.0\%$ & $40.0\%$ & $7.7\%$  & $27.5\%$ \\ 
 \midrule
 RT-1 \citep{rt12022arxiv} &  &\checkmark  &  &$38.0\%$ & $40.0\%$ & $15.4\%$  & $33.8\%$ \\ 
 \midrule
 RT-1 + IL & \checkmark & \checkmark &  &$46.0\%$ & $60.0\%$ & $15.4\%$  & $42.5\%$ \\
 \midrule
 & \checkmark &  & \checkmark &$58.0\%$ & $46.7\%$ & $15.4\%$  & $47.5\%$ \\
 \cmidrule{2-8}
 \textbf{DIAL} (ours)  &  & \checkmark & \checkmark & $50.0\%$ & $37.5\%$ & $10.0\%$ & $36.7\%$\\
 \cmidrule{2-8}
  & \checkmark & \checkmark & \checkmark & $\bm{68.0\%}$ & $\bm{66.7\%}$ & $\bm{30.8\%}$ & $\bm{60.0\%}$ \\
 \bottomrule
\end{tabular}
\caption{
Comparing the performance of language-conditioned policies trained on different types of labeled datasets.
$\mathcal{D}_A$ contains 5,600 episodes with crowd-sourced language instructions and is representative of the Interactive Language (IL) \citep{lynch2022interactive} setting.
$\mathcal{D}_B$ contains 80,000 episodes with structured teleoperator commands and is representative of the RT-1 \citep{rt12022arxiv} setting.
\method additionally creates an augmented $\mathcal{D}_C$ with 38,516 relabeled instructions.
We find that \method is able to significantly performance on novel evaluation instructions, especially in the IL setting where $\mathcal{D}_B$ is not available.}

\label{tab:partial_labels}
\end{center}
\end{table}

\begin{figure*}[t]
    \centering
    \includegraphics[width=0.85\linewidth]{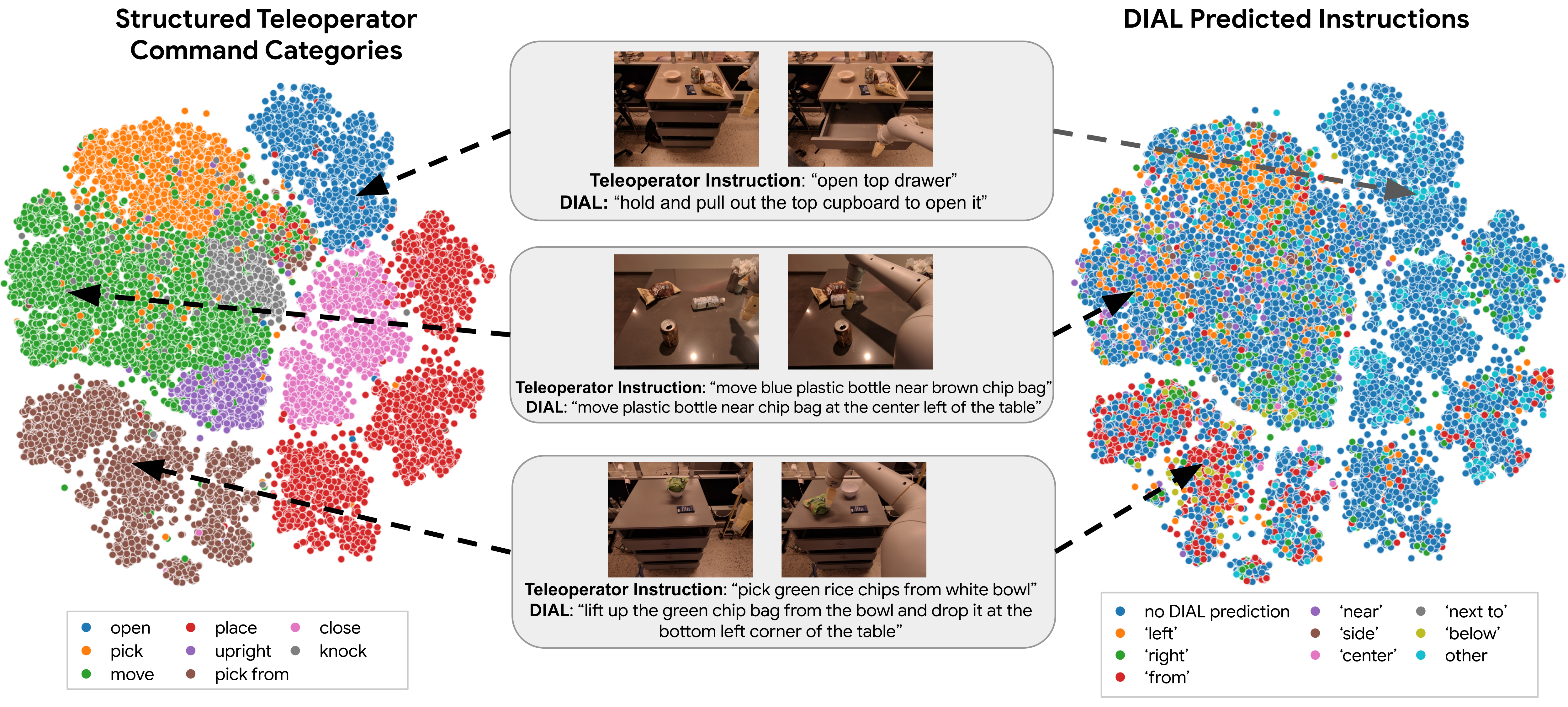}
    \caption{Comparing the task diversity of structured commands provided to teleoperators and DIAL instruction predictions. The visualization shows a t-SNE for 30,000 trajectories with their first and last frames embedded via a fine-tuned CLIP model. On the left, embeddings are colored based on the skill categories of structured teleoperator commands. On the right, embeddings are colored based on particular keywords that are present in the DIAL predicted instructions. While large clusters of episodes may all correspond to the same teleoperator command, DIAL predictions may highlight more nuanced semantic concepts.
    }
    \label{fig:tsne_full}
\end{figure*}

\begin{figure}[h]
    \centering
    \includegraphics[width=0.95\linewidth]{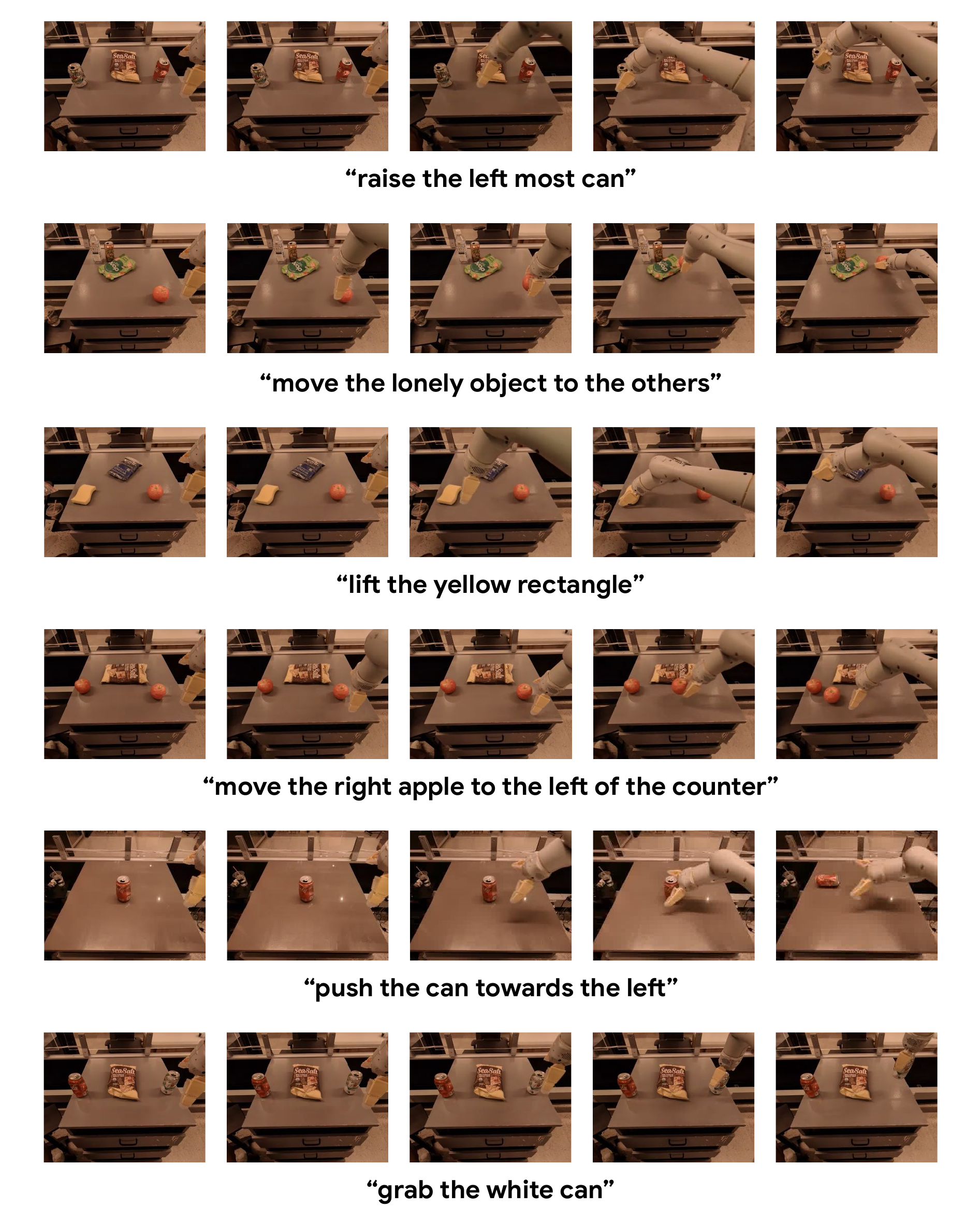}
    \caption{DIAL successfully completes various novel evaluation instructions requiring understanding of concepts such as relative spatial references, colors, and alternative phrasings. These concepts were not present in the structured teleoperator commands used for the original training demonstrations.}
    \label{fig:additional_examples}
\end{figure}

An example is shown in Figure~\ref{fig:qualitative_results}, where DIAL successfully understands the spatial concepts of ``left'', ``middle'', and ``right''.
Such spatial concepts are especially important to identify object instances in scenes with duplicate objects: while baseline methods ignore the language instruction and instead repeat the same motions or randomly select a target object, DIAL is able to consistently target the correct objects.
In addition to \textit{Spatial} tasks, DIAL is also able to outperform baseline policies at \textit{Semantic} tasks that focus on semantic skills not contained in the original foresight instructions.
We show more examples of evaluation successes in Figure~\ref{fig:additional_examples}.

\subsection{How does DIAL compare to other instruction augmentation methods?}
\label{subsec:exp2}
We compare DIAL to non-visual instruction augmentation strategies outlined in Section  \ref{subsec:instruction_augmentation}. 
For this comparison, we apply the baseline instruction augmentation methods on both episodes with crowd-sourced annotations ($\mathcal{D}_A$) and on episodees with structured teleoperator commands ($\mathcal{D}_B$) to produce different instruction augmented datasets ($\mathcal{D}_{C}^{Gaussian}, \mathcal{D}_{C}^{Synonyms}, \mathcal{D}_{C}^{LLM}, \mathcal{D}_{C}^{DIAL}$). Afterwards, we train separate language-conditioned policies policies: the ``None'' model trains on \{$\mathcal{D}_A, \mathcal{D}_{B}$\}, while the remaining models train on \{$\mathcal{D}_A, \mathcal{D}_{B},\mathcal{D}_{C'}$\} with $\mathcal{D}_{C'}$ being each of their respective instruction augmentated datasets. 
Table \ref{tab:main_table} indicates that \method significantly outperforms other baseline instruction augmentation methods. 
For both the \textit{Spatial} Tasks and \textit{Rephrased} tasks, we observe that these baseline instruction augmentation methods without visual grounding resulted in worse performance compared to the no instruction augmentation. 

\begin{table}[t]
\begin{center}
\begin{tabular}{@{}l c c c c@{}} 
 \toprule
  \multicolumn{1}{c}{} & \multicolumn{4}{c}{\textbf{Evaluation on Novel Instructions}} \\
\cmidrule(lr){2-5}
  \textbf{Instruction Augmentation} & \shortstack{\textit{Spatial}\\ Tasks} & \shortstack{\textit{Rephrased} \\ Tasks} & \shortstack{\textit{Semantic} \\ Tasks} & Overall \\ [0.5ex] 
 \midrule
 None & $46.0\%$ & $60.0\%$ & $15.4\%$  & $42.5\%$ \\ 
 \midrule
 Gaussian Noise & $36.0\%$ & $40.0\%$ & $23.1\%$ & $33.8\%$ \\
 \midrule
 Word-level Synonyms & $28.0\%$ & $46.7\%$ & $7.7\%$  & $27.5\%$ \\
 \midrule
 LLM-proposed Instructions  & $28.0\%$ & $46.7\%$ & $23.1\%$ & $30.0\%$ \\
  \midrule
 \textbf{DIAL} (ours)  & $\mathbf{68.0\%}$ & $\mathbf{66.7\%}$ & $\mathbf{30.8\%}$ & $\mathbf{60.0\%}$ \\
 \bottomrule
\end{tabular}
\caption{Evaluating language-conditioned BC policies trained on datasets with different types of instruction augmentation. Each policy performs 80 evaluations over 60 novel task instructions. DIAL is consistently most performant, especially on \textit{Spatial} Tasks requiring visual scene understanding.
}\label{tab:main_table}
\vspace{-7pt}
\end{center}
\end{table}

\subsection{How sensitive is DIAL to hyperparameters and instruction prediction accuracy?}
\label{subsec:accuracy_tradeoff}

\begin{figure}[h]
    \centering
    \includegraphics[width=0.8\linewidth]{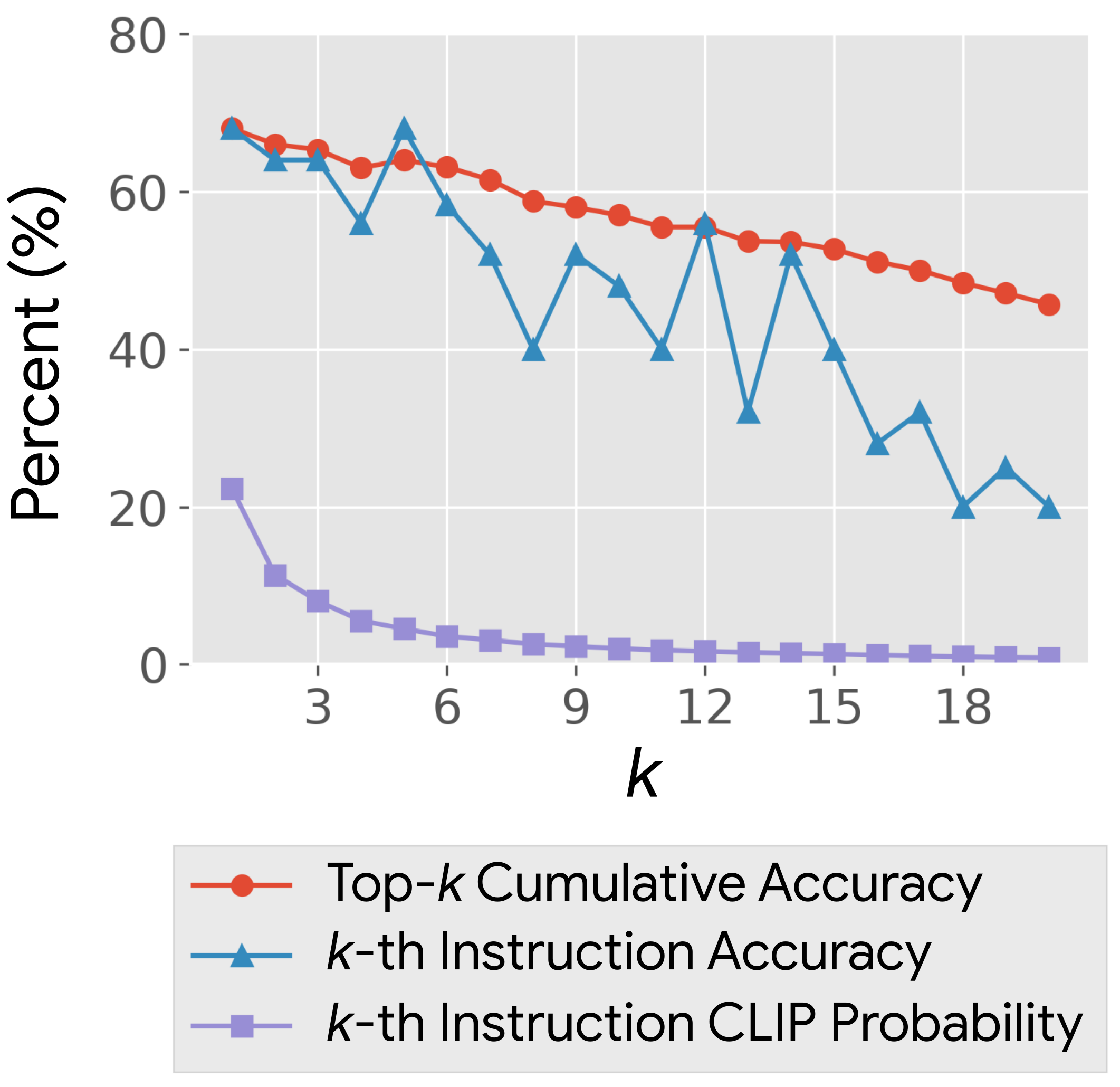}
    \caption{The factual accuracy of the top $20$ instruction augmentation predictions of 50 sampled episodes relabeled by a fine-tuned CLIP model.
    For $k \in [1, 20]$, we measure the accuracy of the $k$-th instruction, the cumulative accuracy of the top $k$ instructions, and CLIP's confidence score for the $k$-th instruction.
    While top instructions are often accurate, they become increasingly inaccurate along with CLIP's confidence.
    }
    \label{fig:instr_accuracy_plot}
\end{figure}

\begin{table*}[t]
\begin{center}
\begin{tabular}{@{}c c c c c c c@{}} 
 \toprule
  \multicolumn{1}{c}{\textbf{DIAL Version}} & \multicolumn{2}{c}{\textbf{Dataset Properties}} & \multicolumn{4}{c}{\textbf{Evaluation on Novel Instructions}} \\
  \cmidrule(lr){1-3} \cmidrule(lr){4-7}
 Prediction Method & \shortstack{Relabeled \\ Episodes} & \shortstack{Relabeled \\ Accuracy} & \shortstack{\textit{Spatial} \\ Tasks} & \shortstack{\textit{Rephrased} \\ Tasks} & \shortstack{\textit{Semantic} \\ Tasks} & Overall \\ [0.5ex] 
 \midrule
 Top-$k$, $k=1$ & $80,000$ & $68.0\%$ & $62.0\%$ & $40.0\%$ & $23.1\%$  & $50.0\%$ \\ 
 \midrule
 Top-$k$, $k=3$ & $240,000$ & $65.3\%$ & $62.0\%$ & $40.0\%$ & $15.4\%$  & $48.8\%$ \\ 
 \midrule
 Top-$k$, $k=10$ & $800,000$ & $57.0\%$  & $37.5\%$ & $50.0\%$ & $20.0\%$ & $35.0\%$\\
 \midrule
 Min-$p$, $p=0.10$ & $128,422$ & $61.9\%$ & $44.0\%$ & $46.7\%$ & $23.1\%$  & $40.0\%$ \\
  \midrule
 Min-$p$, $p=0.20$ & $38,516$ & $68.8\%$ & $\bm{68.0\%}$ & $\bm{66.7\%}$ & $\bm{30.8}\%$ & $\bm{60.0\%}$ \\
  \midrule
 Min-$p$, $p=0.30$ & $17,013$ & $76.0\%$ & $62.0\%$ & $53.3\%$ & $46.2\%$ & $56.3\%$ \\
 \bottomrule
\end{tabular}
\caption{Comparing DIAL with Top-$k$ prediction against DIAL with Min-$p$ prediction.
By increasing $k$ or decreasing $p$, augmented datasets become larger but increasingly inaccurate.
We provide analysis of the relationship between instruction accuracy and CLIP confidence in Figure~\ref{fig:instr_accuracy_plot} and Section~\ref{subsec:accuracy_tradeoff}.
}\label{tab:dial_ablation}
\end{center}
\end{table*}

We study the trade-off between increasing the amount of instruction augmentation and potentially relabeling with incorrect or irrelevant instructions.
By varying the hyperparameters of Top-$k$ prediction and Min-$p$ prediction, the two instruction prediction variations of DIAL discussed in Section~\ref{subsec:instruction_augmentation}, we can indirectly influence the size the potential label inaccuracy of the relabeled datasets.
To measure how instruction augmentation accuracy changes as we increase \textit{k}, we ask human labelers to rate whether proposed instruction augmentation are factually accurate descriptions of a given episode.
We show an example of the top 10 predicted instruction augmentations in an episode in Figure~\ref{fig:instruction_accuracy_examples}.

In Figure \ref{fig:instr_accuracy_plot}, we sample 50 episodes and ask human labelers to assess the predicted instruction accuracy as we increase the number of predictions produced by CLIP. While the initial predictions are often correct, the later predictions are often factually inaccurate. The top-20th instruction prediction is only factually accurate $20.0\%$ of the time. 

When applying these different relabeled datasets to downstream policy learning, we find in Table~\ref{tab:dial_ablation} that Min-$p$ instruction prediction, a more conservative approach than Top-$k$ prediction, performs the best across all evaluation instructions.
We also find that finetuning CLIP is quite important for instruction augmentation, which is detailed in Appendix \ref{app:finetuning}.

%% file: 6_conclusion.tex
\section{Conclusion, Limitations, and Future Work}
In this work, we introduced \method, a method that uses VLMs to label offline datasets for language-conditioned policy learning.
Scaling \method to a real world robotic manipulation domain, we perform a large-scale study of over 1,300 evaluations and find that DIAL is able to outperform baselines by 40\% on a challenging set of 60 novel evaluation instructions unseen during training.
We compare DIAL against instruction augmentation methods that do not consider visual context, and also ablate the source datasets we use for instruction augmentation.
Finally, we study the interplay between larger augmented datasets and lowered instruction accuracy; we find that control policies are able to utilize relabeled demonstrations even when some labels are inaccurate, suggesting that DIAL is able to provide a cheap and automated option to extract additional semantic knowledge from offline control datasets.

\paragraph*{Limitations and Future Work} Although DIAL seems to improve policy understanding on many novel concepts not contained in the original training dataset, it sometimes fails, especially when evaluating tasks that may require new motor skills.
In addition, since both our crowd-source annotators and VLMs only have access to the first and final states of an episode, they do not capture skills involving temporal coherence nor is aware of \textit{how} these instructions are accomplished.
A natural next step is to apply DIAL to full video episodes. 
Another interesting direction is to view DIAL as goal-conditioning and attempting visual goals during training or evaluation.
Moreover, on-policy or RL variations of DIAL may be able to effectively explore the task representation space autonomously.

%% file: 7_appendix.tex
\subsection{Additional Experiments}
We present failure examples, study the importance of both pre-training and fine-tuning, and also explore whether VLMs used for DIAL could also be utilized as a task representation.

\subsubsection{Failure Examples}
\begin{figure}[h]
    \centering
    \includegraphics[width=0.95\linewidth]{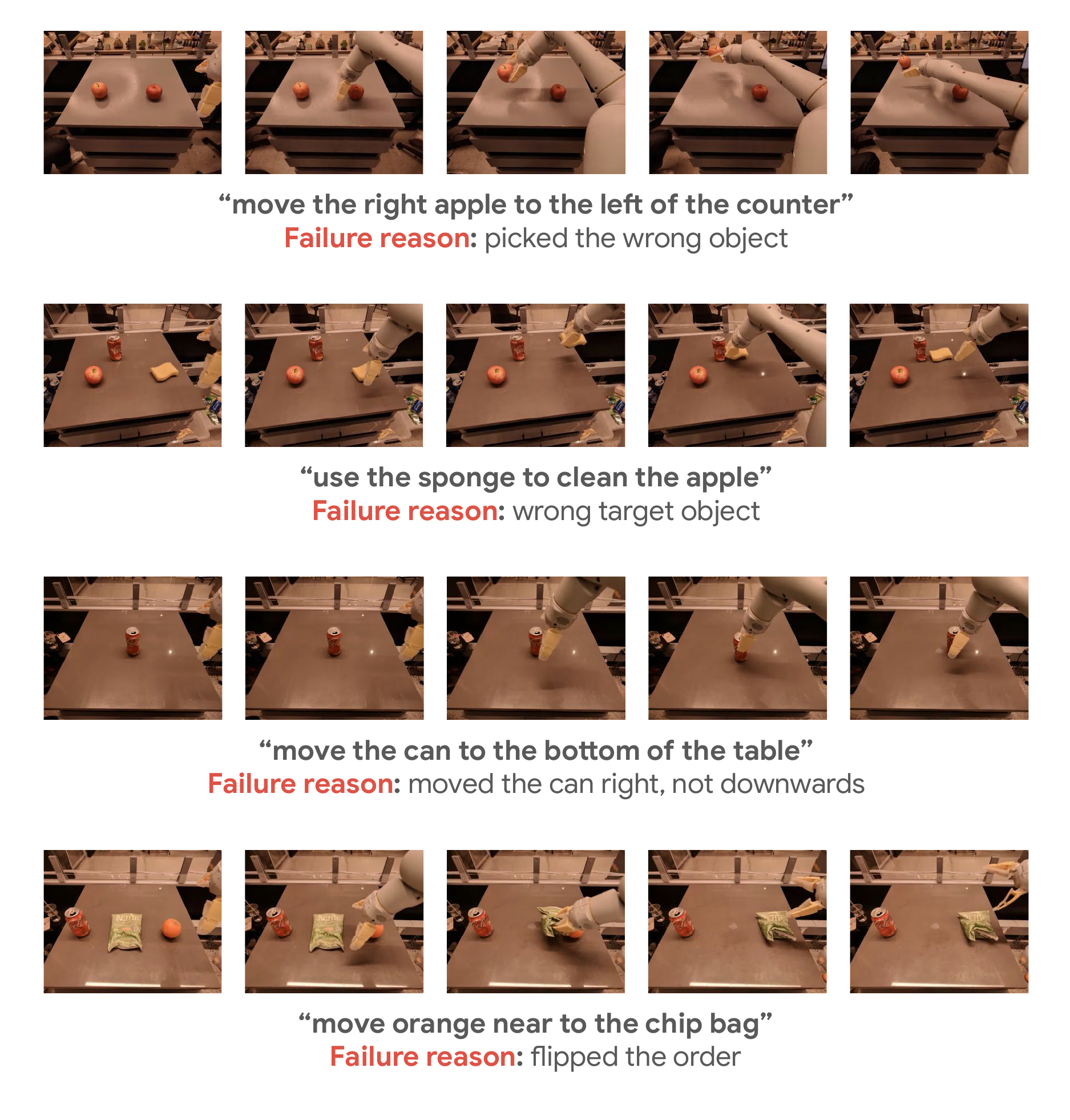}
    \caption{Samples of evaluation failures for DIAL. Errors are due to a combination of motor control and task confusion.}
    \label{fig:failure_examples}
\end{figure}

In addition to the successful example trajectories visualized in Figure~\ref{fig:qualitative_results} and Figure~\ref{fig:additional_examples}, we also show some examples of failure cases in Figure~\ref{fig:failure_examples}.

\subsubsection{How important is VLM fine-tuning for DIAL?}
\label{app:finetuning}

Whereas Section~\ref{subsec:accuracy_tradeoff} studies different prediction mechanisms for a fine-tuned CLIP model (FT-CLIP), we are also interested in comparing different CLIP models altogether.
The main FT-CLIP model used in DIAL is initialized from the pretrained OpenAI CLIP weights and then fine-tuned on $\mathcal{D}_A$ from Section~\ref{subsec:finetuning}, which begs the question: are both a strong pretrained initialization and subsequent fine-tuning necessary for strong instruction labeling performance?
To answer this question, we perform instruction augmentation with (1) the frozen pretrained OpenAI CLIP model and (2) a fine-tuned CLIP model that starts from a random weight initialization instead of from the OpenAI pretrained weights.

\begin{figure}[h]
    \centering
    \includegraphics[width=0.90\linewidth]{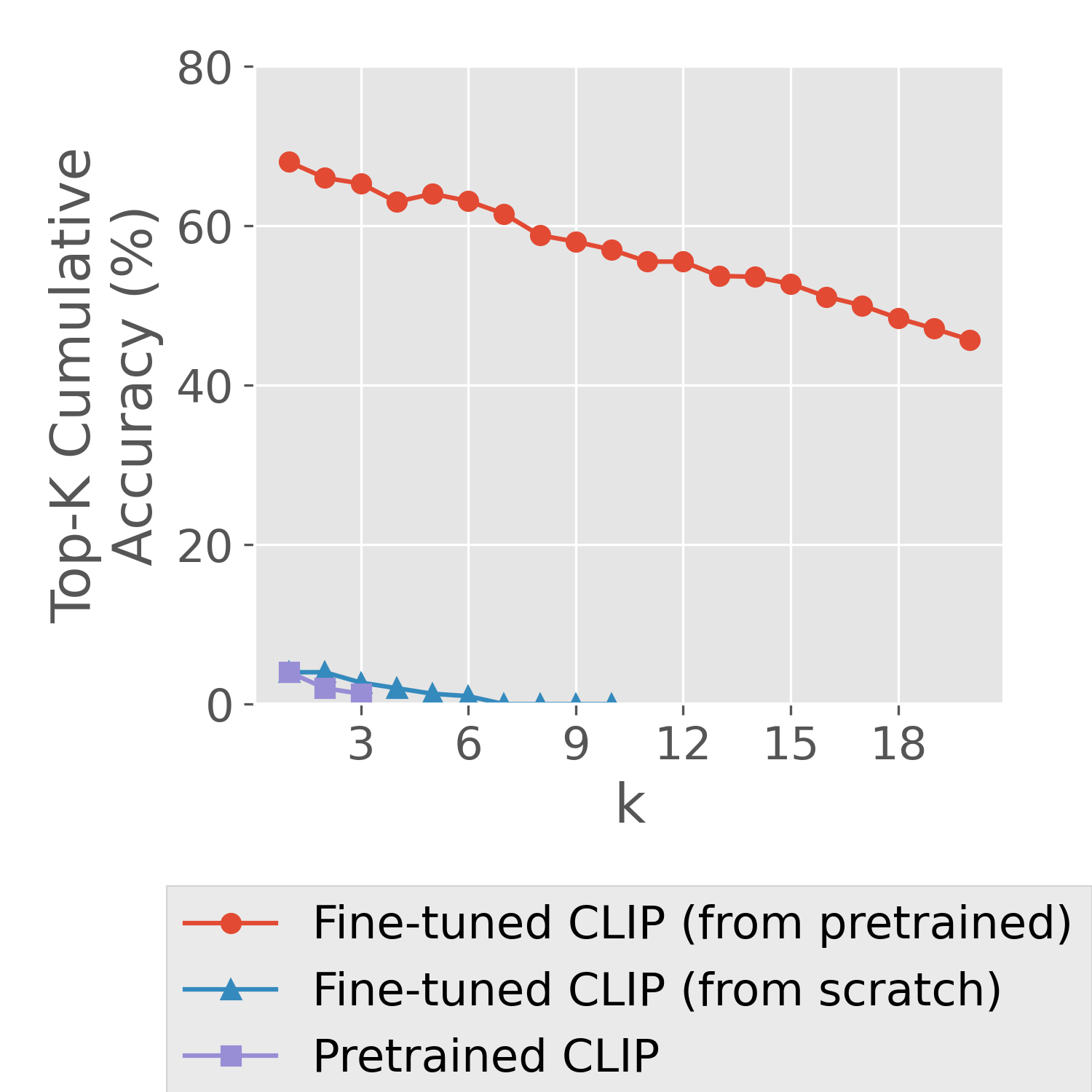}
    \caption{Estimated top-$K$ cumulative instruction labeling accuracy using CLIP variants. The fine-tuned CLIP model from OpenAI checkpoint \citep{radford2021learning} performed significantly better than frozen CLIP model and the fine-tuned model from random initial weights.
    }
    \label{fig:clip_finetuning_comparison}
\end{figure}

\begin{figure}[h]
    \centering
    \includegraphics[width=0.85\linewidth]{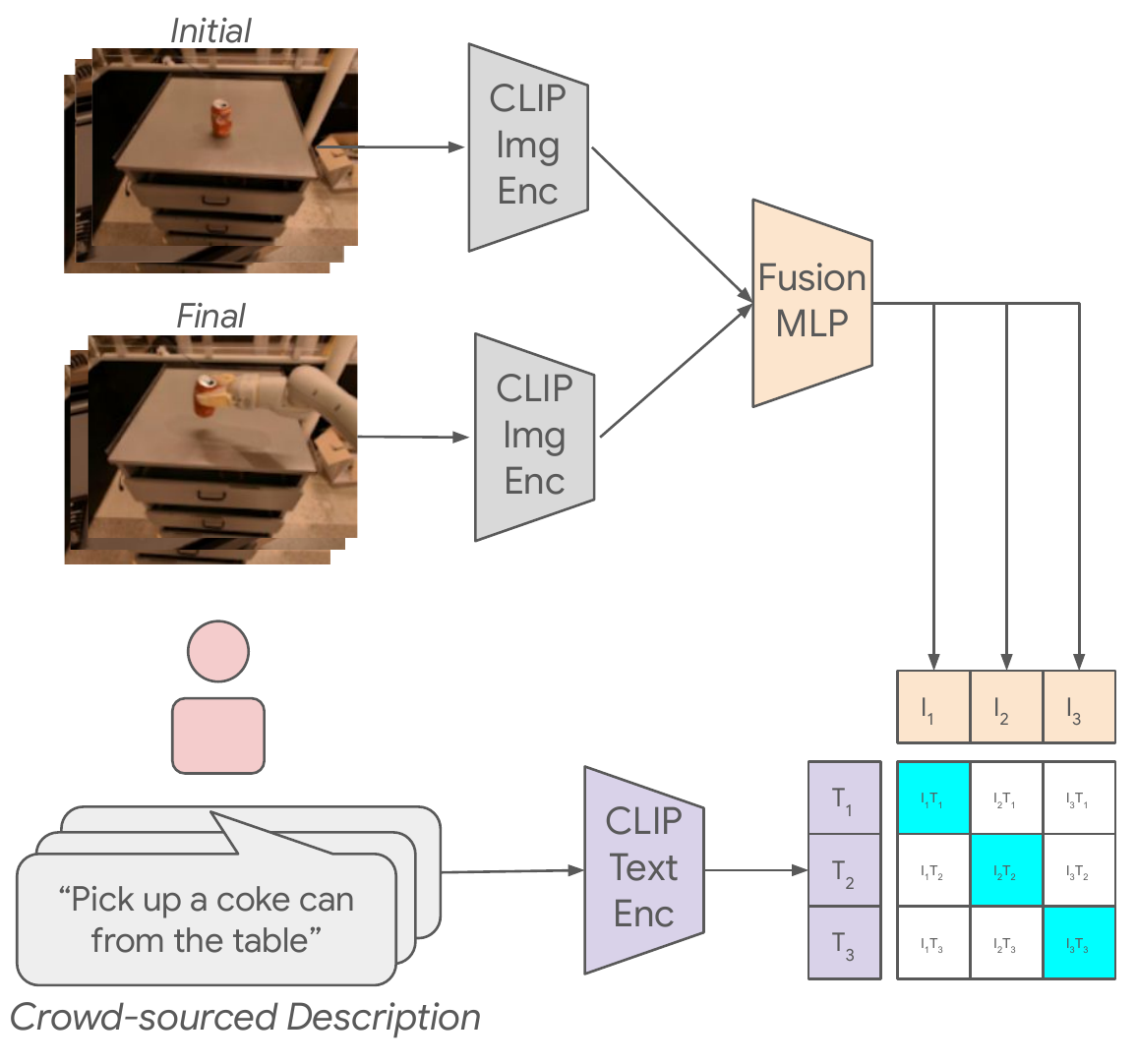}
    \caption{CLIP architecture for fine-tuning using the contrastive loss. The image embeddings of the initial and final observation is concatenated and passed through an MLP to produce an episode embedding to dot product with the text embedding.}
    \label{fig:clip_arch}
\end{figure}

As we see in Figure \ref{fig:clip_finetuning_comparison}, predicted instruction accuracy for both of these models is significantly lower than the main FT-CLIP model.
The poor performance of the frozen pretrained CLIP model (1) suggests that the particular embodied captioning task required by DIAL is likely quite out of distribution for the pre-training used by the OpenAI CLIP model, so some amount of domain data is required.
On the other hand, the poor performance of training CLIP from scratch on robot demonstration data (2) suggests that internet-scale pre-training is required to start from a reasonable prior; there may not be sufficient robot domain data in the datasets we consider to fully train CLIP from scratch.

\subsubsection{Is a VLM good at relabeling also a good task representation?}
\label{app:task_representation}
We study whether a VLM fine-tuned for instruction augmentation can also act as a better task representation for conditioning a policy in the form of a more powerful language embedding.
Across the various groundtruth and relabeled datasets we focus on, we find that fine-tuned CLIP (FT-CLIP) is the most effective task representation, as seen in Table~\ref{tab:clip_representation}.
FT-CLIP is a good representation not only for freeform language instructions like those contained in the fine-tuning dataset $\mathcal{D}_A$, but also for structured metadata labels used to collect the demonstrations in $\mathcal{D}_B$.
Thus, we utilize FT-CLIP's language encoder as the main task representation for language conditioning in all control policies we train, besides for policies that explicitly denote otherwise, such as in Table~\ref{tab:clip_representation}.

\begin{table}[h]
\begin{center}
\begin{tabular}{@{}c c p{0.7cm} p{0.9cm} p{0.7cm} p{0.5cm}@{}} 
 \toprule
 & & \multicolumn{4}{c}{\textbf{Evaluation on Novel Instructions}} \\
 \cmidrule(lr){3-6}
 Dataset & Task Encoder & \textit{Spatial} & \textit{Rephrased} & \textit{Semantic} & Overall \\ [0.5ex] 
 \midrule
$\mathcal{D}_A$ & USE & $22.5\%$ & $50.0\%$ & $0.0\%$  & $21.7\%$ \\  
 \midrule
$\mathcal{D}_A$ & FT-CLIP  & $30.0\%$ & $40.0\%$ & $7.7\%$  & $27.5\%$ \\ 
 \midrule
$\mathcal{D}_A$, $\mathcal{D}_B$ & CLIP  & $45.0\%$ & $40.0\%$ & $10.0\%$  & $40.0\%$ \\ 
 \midrule
$\mathcal{D}_A$, $\mathcal{D}_B$ & FT-CLIP & $46.0\%$ & $60.0\%$ & $15.4\%$  & $42.5\%$ \\
  \midrule
DIAL, $k=1$ & USE & $50.0\%$ & $50.0\%$ & $20.0\%$  & $43.3\%$ \\ 
  \midrule
DIAL, $k=1$ & FT-CLIP & $62.0\%$ & $40.0\%$ & $23.1\%$  & $50.0\%$ \\
 \bottomrule
\end{tabular}
\caption{Comparing downstream policy performance when improving the task representation from USE~\citep{cer2018universal} to Pretrained OpenAI CLIP (CLIP)~\citep{radford2021learning} to fine-tuned CLIP (FT-CLIP), as described in Section \ref{subsec:finetuning}. We find that the FT-CLIP representation is the best task representation in all dataset settings: training on crowd-sourced language annotations ($\mathcal{D}_A$), training on structured teleoperator commands along with crowd-sourced language annotations \{$\mathcal{D}_A$, $\mathcal{D}_B$\}, and using DIAL \{$\mathcal{D}_A$,$\mathcal{D}_B$,$\mathcal{D}_C$\} with Top-$k$ with $k = 1$ (DIAL, $k=1$).
}\label{tab:clip_representation}
\end{center}
\vspace{-4mm}
\end{table}

\subsection{Dataset Details}
\label{app:datasets}
Following the procedure in \citep{rt12022arxiv}, we collect a large dataset of robot trajectories via teleoperation by uniformly sampling from one of the structured commands shown in Table~\ref{tab:dataset_skills} and sending those commands to teleoperators that operate a mobilee manipulator robot in the real world.
After teleoperators discard unsafe or failed demonstrations, we save a dataset of 80,000 successful robot demonstration trajectories ($\mathcal{D}_B$).
Then, we send 2,800 of the episodes from $\mathcal{D}_B$ to be labeled by a pool of human labelers, who see the first and last frame of the episode and are tasked with providing a natural language description of how a robot could be commanded to from the first frame to the last frame. 
We request two independent language instruction annotations for each episode, so we obtain a total of 5,600 episode-instruction pairs, which we save as $\mathcal{D}_A$.
Finally, we produce various relabeled datasets via instruction augmentation, that we detail in Section~\ref{subsec:instruction_augmentation} for different prediction methods for \method and Section~\ref{subsec:exp2} for different non-visual instruction augmentation baselines.

\begin{table*}[t]
\begin{center}
\begin{tabular}{p{4.5cm} p{0.75cm} p{4.5cm} p{4.5cm}}
\toprule
Structured Teleoperator Command Categories & Count & Description & Example Instruction\\
\midrule

Pick \texttt{Object} & 17 & picking objects on a counter & pick water bottle \\

Move \texttt{Object} Near \texttt{Object} & 342 & moving an object near another & move pepsi can near rxbar blueberry\\

Place \texttt{Object} Upright & 8 & placing an elongated object vertically upright & place coke can upright\\

Knock \texttt{Object} Over & 8 & picking an object and laying it sideways on the counter & knock redbull can over\\

Open / Close \texttt{Drawer} & 6 & opening or closing a counter drawer & open the top drawer \\
Place \texttt{Object} into \texttt{Receptacle} & 85 & pick an object on the table and put it in a container or drawer & place brown chip bag into white bowl\\
Pick \texttt{Object} from \texttt{Receptacle} and Place on the Counter & 85 & pick an object out of a container or drawer and place it on the counter & pick green jalapeno chip bag from paper bowl and place on counter \\
\midrule
Total & 551 \\
\bottomrule
\end{tabular}
\caption{Teleoperators receive an instruction sampled from a total of 551 unique structured commands covering 7 different manipulation skills. The 80,000 episodes in $\mathcal{D}_B$ each contain exactly one of these structured teleoperator commands.
}\label{tab:dataset_skills}
\end{center}
\end{table*}

\subsection{DIAL Implementation Details}
\label{app:dial_implementation}

\begin{figure*}[t]
    \centering
    \includegraphics[width=0.95\linewidth]{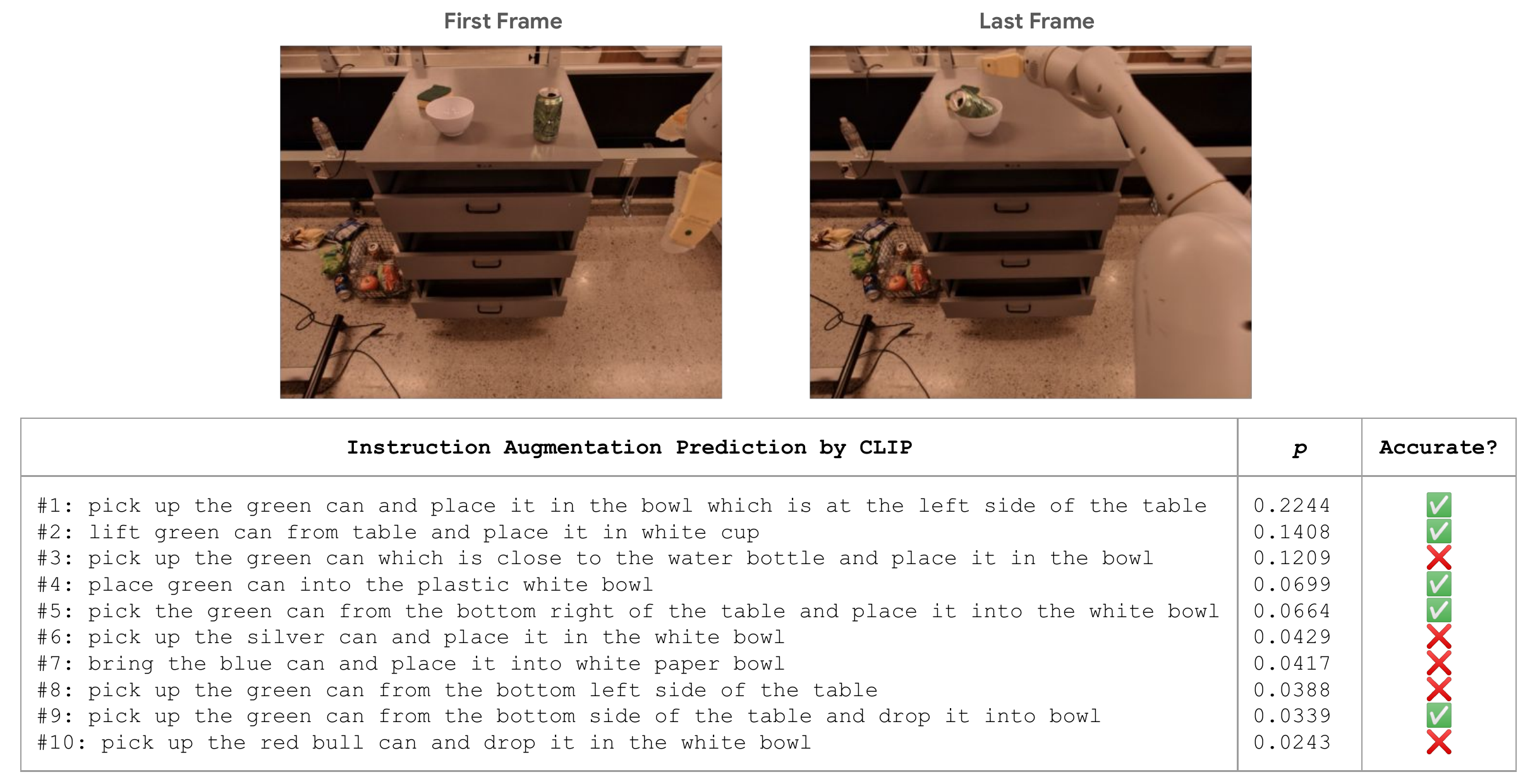}
    \caption{The top 10 proposed instruction augmentations for a single episode with original structured teleoperator command \texttt{place green can in white bowl}. In some cases, the predicted captions provide additional semantic information such as describing the location of the can or the material of the bowl. As seen in Figure~\ref{fig:instr_accuracy_plot}, the probability CLIP assigns to each the candidates quickly drops off past a few top predictions. Setting Min-$p=0.2$ would only take the first instruction augmentation prediction, while setting $k=3$ would take the top three predictions, including an incorrect prediction (\#3). }
    \label{fig:instruction_accuracy_examples}
\end{figure*}

We implement DIAL with a CLIP model that is fine-tuned on 5,600 annotated episodes ($\mathcal{D}_A$) with the procedure described in Section~\ref{subsec:finetuning}.
The architecture of the CLIP model is shown in Figure \ref{fig:clip_arch}.
The initial and final state observation images are embedded using the CLIP image encoder $I_{enc}$, and the resulting embeddings are concatenated and passed through a 200 hidden dimension single-layer MLP to produce the final episode embedding.
We use the CLIP text encoder $T_{enc}$ to embed the crowd-sourced annotations to produce the corresponding text embeddings.
To fine-tune, we loaded the CLIP encoder weights from the \texttt{ViT-B/32} OpenAI checkpoint. 
We use a batch size of 64 and train for 100,000 iterations, but select the best checkpoint based on text prediction accuracy on a randomly held-out test set of $10\%$ of the training dataset $\mathcal{D}_A$.
We fine-tune both encoders $I_{enc}$ and $T_{enc}$ as well as the fusion MLP. 

After fine-tuning CLIP, we source 18,719 candidate instruction labels ($L$) from $\mathcal{D}_A$ and a corpus of GPT-3 proposals of potential language instructions.
The GPT-3 proposals are generated by using the prompt shown in Listing~\ref{lst:prompt_candidates} to iterate over the 551 instructions used to collect teleoperated demonstrations.
We note that that Listing~\ref{lst:prompt_candidates} generates diverse instructions that may not be accurate for a given episode.
Listing~\ref{lst:prompt_candidates} is purposefully tuned to produce ``hallucinated'' descriptions that can add semantic properties in the proposed instructions that may or not be correct (for example, ``pick up the orange'' might be augmented into ``retrieve the orange from the sink'' or ``raise the orange next to the vase'').
The motivation behind this design decision is that GPT-3 predictions can be a lot less conservative when being used downstream by DIAL, since the CLIP model will ideally filter out irrelevant instructions.
In contrast, the prompt in Listing~\ref{lst:prompt_synonyms} is used for producing Sentence-Level Synonyms, which should ideally always be factually equivalent to the original instruction.

Next, to relabel 80,000 robot trajectories that do not contain crowd-sourced annotations ($\mathcal{D}_B$) with $L$, we follow Section~\ref{subsec:relabeling} to implement two variations of DIAL: \textbf{Top-$k$ selection} and \textbf{Min-$p$ selection}.
For these two variations, we use $k=\{1, 3, 10\}$ and $p = \{0.1, 0.2, 0.3\}$.

\subsection{DIAL Hyperparameters}
In Section \ref{subsec:accuracy_tradeoff}, we examined DIAL's sensitivity to hyperparameters, including top-K and min-P sampling, which influence dataset diversity and instruction accuracy (label noise). Such tradeoffs are common in data augmentation, where the benefits of increased augmentation must be balanced with the risks of too much label noise. These hyperparameter choices are highly domain-specific, and in our case we optimize this balance by selecting hyperparameters based on measuring offline VLM prediction accuracy. Notably, hyperparameter tuning in DIAL is substantially more cost-effective than manual full dataset labeling by humans, enabling scalable policy learning improvement from generated instructions. We plan to include further discussion on hyperparameter selection and future work on studying these tradeoffs in more detail.

\subsection{Instruction Augmentation Baselines}
\label{app:baseline_implementations}

\paragraph{Gaussian Noise}
Given an instruction $l$, we add Gaussian noise to the language embedding produced by the CLIP text encoder $T_{enc}$, directly obtaining the augmentation in the latent space $\tilde{z}_l$:
\begin{align}
    \tilde{z}_{l} = T_{enc}(l) + \epsilon, \quad \epsilon \sim \mathcal{N}(0,\sigma) \in \mathbb{R}^d
\end{align}

In our implementation, we choose $\sigma=0.05$ and perform the Gaussian noise augmentation dynamically to the 512-dimension CLIP $T_{enc}$ embedding resulting from passing in the original language instruction to the CLIP text encoder.

\paragraph{Word-level Synonyms}
We replace \textit{individual words} in existing instructions with sampled synonyms from a predefined list.
The mapping between words present in the original structured 551 instructions and possible synonyms is shown in Listing~\ref{lst:word_synonyms}.

\paragraph{Sentence-level Synonyms} 
We replace entire instructions with alternative instructions as proposed by GPT-3.
We pre-compute valid sentence-level synonyms by using the prompt shown in Listing~\ref{lst:prompt_candidates} to iterate over the 551 instructions used to collect teleoperated demonstrations.

\subsection{Augmented Dataset Details}
As noted in Section \ref{subsec:instruction_augmentation}, DIAL uses CLIP to score predictions from 18,719 candidate text labels $L$, which are a union of 9,393 crowd-sourced instructions ($L_{CC}$) from the original $\mathcal{D}_A$ dataset as well as 9,336 instruction candidates ($L_{GPT3}$) from GPT-3 curated with the prompt in Listing 1.
During the instruction augmentation process, we find that CLIP selects 3,675 unique instructions from these 18,719 candidates $L$. Additionally, $86.3\%$ of selected instructions were from $L_{CC}$ and $13.7\%$ from $L_{GPT3}$. These characteristics show that CLIP generally prefers human-like instructions while sometimes incorporating GPT-3 variations.
We note that these prediction characteristics may result from the fact that our relabeling model CLIP was finetuned solely on crowd-sourced labels, and that the desired output distribution is unclear (is it a positive or negative property to prefer crowd-sourced annotations?).
Furthermore, it's possible that the true distribution of accurate and desireable language descriptions was already well-covered by the human annotations; so if GPT-3 proposed an instruction already contained in the human-sourced $L_{CC}$, we would deduplicate that candidate from $L$ and consider the instruction to be human-sourced.
To enhance diversity, future work could leverage improved VLMs and fine-tune on more diverse labels, including LLM-generated captions.

\subsection{Language-Conditioned Policy Training}
\label{app:lcbc}

The policies used in this work are trained using the RT-1~\citep{rt12022arxiv} architecture on a large dataset of human-provided demonstrations.
The policies receive natural language descriptions in the form of a 512-dimensional VLM embedding and a short history of images and outputs discrete action tokens which are then transformed to continuous action outputs. 
Apart from the experiments explicitly denoted as using USE~\citep{cer2018universal} and frozen CLIP~\citep{radford2021learning} language embeddings in Table~\ref{tab:clip_representation}, all policies trained in this work utilize fine-tuned CLIP (FT-CLIP) language embeddings as described in Section \ref{subsec:finetuning}.

Note that the exact policy architecture is not the main focus of this work, so we utilize the exact same policy training procedure across each experiment and only vary the instruction augmented datasets that the policies are trained on.

\subsection{Evaluation Instructions}
We utilize an evaluation setup focusing solely on novel instructions unseen during training.
To source these novel instructions, we 1) crowd-source instructions from a different set of humans than the original dataset labelers and 2) prompt GPT-3 with Listing~\ref{lst:prompt_synonyms} to produce reasonable tasks that might be asked of a home robot manipulating various objects on a kitchen counter.
Then, we normalize all instructions by removing punctuation, removing non-alphanumeric symbols, converting all instructions to lower case, and removing leading and ending spaces.
Afterwards, we filter out any instructions already contained in either the instruction augmentation process in Section~\ref{subsec:relabeling} or in the original set of 551 foresight tasks in Table~\ref{tab:dataset_skills}.
Finally, as seen in Table~\ref{tab:full_eval_tasks}, we organize them into various semantic categories to allow for more detailed analysis of quantitative policy performance.

\begin{minipage}{\linewidth}
\begin{lstlisting}[basicstyle=\ttfamily\scriptsize, backgroundcolor = \color{lightgray}, keywords = {Human, Robot},  keywordstyle = {\textbf}, caption={GPT-3 Prompt for Proposing Candidate Tasks.}, label={lst:prompt_candidates}]
For the following tasks for a helpful home robot, rephrase 
them to imagine different variations of the task. These 
variations include different types of objects, different 
locations, different obstacles, and different strategies
for how the task should be accomplished.

3 rephrases for: pick mountain dew
Answer: lift the mountain dew on the left side of the desk, 
grab the mountain dew soda next to the water, pick the 
farthest green soda can

4 rephrases for: move your arm to the right side of the desk
Answer: bring your arm to the right of the counter, move 
right slightly, go far to the rightmost part of the table, 
reorient your gripper to point right

10 rephrases for: bring me the yogurt
Answer: retrieve the yogurt, bring the white snack, pick up 
the yogurt cup from the far right, lift the yogurt snack 
from the left, bring back the yogurt near the chip bag, lift
the yogurt from the top of the counter, bring the yogurt 
closest to the apple, grab the yogurt, lift the close left 
yogurt on the bottom left, retrieve the yogurt on the 
bottom of the table

10 rephrases for: <INSTRUCTION_TO_AUGMENT>
Answer:
\end{lstlisting}
\end{minipage}

\begin{minipage}{\linewidth}

\begin{lstlisting}[basicstyle=\ttfamily\scriptsize, backgroundcolor = \color{lightgray}, keywords = {Human, Robot},  keywordstyle = {\textbf}, caption={GPT-3 Prompt for ``Sentence-level Synonyms''.}, label={lst:prompt_synonyms}]
You are a helpful home robot in an office kitchen. You are 
able to manipulate household objects in a safe and efficient 
manner. Here are some tasks you are able to accomplish in 
various environments:

5 tasks in a sink with a sponge, brush, plate, and a cup:
move sponge near the cup, fill up the cup with water, clean
the plate with the brush, pick up the plate, put the cup 
on the plate

3 tasks in a storage room with a box, a ladder, and a 
hammer: lift the hammer, push the ladder, put the hammer
in the box

10 tasks on a table with an apple, a coke can, a sponge, and 
an orange: pick up the apple, pick up the coke can, use the 
sponge to clean the apple, use the sponge to clean the coke 
can, put the apple down, put the coke can down, pick up the
orange, peel the orange, eat the orange, throw away the peel

10 tasks on a table with <OBJECT_1>, <OBJECT_2>, and
<OBJECT_3>:

\end{lstlisting}
\end{minipage}

\begin{minipage}{\linewidth}

\begin{lstlisting}[basicstyle=\ttfamily\scriptsize, backgroundcolor = \color{lightgray}, keywords = {Human, Robot},  keywordstyle = {\textbf}, caption={Synonym Mapping for ``Word-level Synonyms''.}, label={lst:word_synonyms}]
SYNONYM_MAP = {
    'rxbar blueberry': [
        'rxbar blueberry', 'blueberry rxbar', 
        'the blueberry rxbar', 'the rxbar blueberry'
    ],
    'rxbar chocolate': [
        'rxbar chocolate', 'chocolate rxbar', 
        'the chocolate rxbar', 'the rxbar chocolate'
    ],
    'pick': ['pick', 'pick up', 'raise', 'lift'],
    'move': [
        'move', 'push', 'move', 'displace', 'guide', 
        'manipulate', 'bring'
    ],
    'knock': ['knock', 'push over', 'flick', 'knockdown'],
    'place': ['place', 'put', 'gently place', 'gently put'],
    'open': ['open', 'widen', 'pull', 'widely open'],
    'close': ['close', 'push close', 'completely close'],
    'coke': [
        'coke', 'coca cola', 'coke', 'coca cola', 
        'the coke', 'a coke', 'a coca cola', 'the coca cola'
    ],
    'green': [
        'green', 'bright green', 'grass colored', 'lime', 
        'a green', 'the green', 'a lime', 'the lime',
        'the bright green', 'a bright green'
    ],
    'blue ': ['blue ', 'dark blue ', 'the blue ', 'a blue '],
    'pepsi': ['pepsi', 'blue pepsi', 'pepsi', 'a pepsi',
        'the pepsi'],
    '7up': ['7up', 'white 7up', '7up', '7-up', '7up', 
        'a 7up', 'the 7up'],
    'redbull': [
        'redbull', 'red bull', 'energy drink', 
        'redbull energy', 'redbull soda',
        'the redbull', 'a redbull', 'a red bull', 
        'the red bull'
    ],
    'blueberry': ['blueberry', 'blue berry'],
    'chocolate': ['chocolate', 'brown chocolate'],
    'brown': ['brown', 'coffee colored', 
                'the brown', 'a brown'],
    'jalapeno': ['jalapeno', 'spicy', 'hot', 'fiery'],
    'rice': ['rice'],
    'chip': ['chip', 'snack', 'chips'],
    'plastic': ['plastic'],
    'water': ['water', 'water', 'agua'],
    'bowl': ['bowl', 'half dome', 'chalice'],
    'togo': ['togo', 'to-go', 'to go'],
    'box': ['box', 'container', 'paper box'],
    'upright': ['upright', 'right side up', 'correctly'],
    'near': ['near', 'close to', 'nearby', 
            'very near', 'very close to'],
    'can': ['can', 'soda can', 'aluminum can'],
    'rxbar': ['rxbar', 'snack bar', 'granola bar',
        'health bar', 'granola'],
    'apple': [
        'apple', 'red apple', 'the apple', 'the red apple', 
        'an apple', 'a red apple', 'small apple', 
        'the small apple'
    ],
    'orange': [
        'orange', 'the orange', 'orange fruit', 'an orange',
        'a small orange', 'a large orange'
    ],
    'sponge': [
        'sponge', 'yellow sponge', 'the yellow sponge', 
        'a yellow sponge', 'a sponge', 'the sponge'
    ],
    'bottle': ['bottle', 'plastic bottle',
                'recycleable', 'clear'],
}

\end{lstlisting}
\end{minipage}

\begin{table*}[t]
\small
\small
\begin{center}
\renewcommand{\arraystretch}{1.2}
         \centering
\begin{tabular}{|p{0.1\textwidth}|p{0.8\textwidth}|}
\hline
\textbf{Category} & \textbf{Instruction Samples}\\ \hline
\textit{Spatial} & \texttt{[`grab the bottle on the left of the table',
`grab the can which is on the right side of the table',
`grab the chip on the left',
`grab the chip on the right',
`grab the right most apple',
`knock down the right soda',
`lift the apple which is on the left side of the table',
`lift the apple which is on the right side of the table',
`lift the chips on the left side',
`lift the chips on the right side',
`lift the left can',
`move the left soda to the can on the right side of the table',
`move the soda can which is on the right toward the chip bag',
`pick can which is on the left of the table',
`pick can which is on the right of the table',
`pick chip bag on the left',
`pick chip bag on the right',
`pick the can in the middle',
`pick the left coke can',
`pick the left fruit',
`pick the leftmost chip bag',
`pick the object on the right side of the table',
`pick the right coke can',
`pick the right object',
`pick the rightmost chip bag',
`pick up the left apple',
`pick up the left object',
`pick up the right can',
`pick up the right object',
`push the left side apple to the brown chips',
`raise bottle which is to the left of the can',
`raise the blue tin',
`raise the left most can',
`raise the thing which is on the left of the counter']} \\ \hline
\textit{Rephrased} & \texttt{ [`grab and lift up the green bag',
`grab the blue pepsi',
`grab the white can',
`knock over the water',
`lift the orange soda',
`lift the yellow rectangle',
`liftt the fruit',
`move green packet near the red apple',
`move orange near to the chip bag',
`pick up the apple fruit',
`push green chips close to the coke',
`upright the lime green can',
`put the apple next to the candy bar',
`retrieve the can from the left side of the coffee table',
`set the apple down next to the chocolate bar',
`take the can from the left side of the counter']} \\ \hline
\textit{Semantic} & \texttt{[`move the can to the bottom of the table',
`move the green bag away from the others',
`move the lonely object to the others',
`move the right apple to the left of the counter',
`push blue chip bag to the left side of the table',
`push the can towards the left',
`push the can towards the right',
`push the left apple to the right side',
`use the sponge to clean the coke can',
`use the sponge to clean the apple']} \\ \hline
\end{tabular}
\caption{Novel evaluation instructions sourced from humans or GPT-3, grouped by category. Spatial tasks focus on tasks involving Spatial relationships, Rephrased tasks contain tasks that directly map to a foresight skill, and Semantic tasks describe semantic concepts not contained in the relabeled or original datasets. In total, there are 60 instructions across the three categories.}
\label{tab:full_eval_tasks}
\end{center}
\end{table*}